\documentclass[10pt,twocolumn,letterpaper]{article}

\usepackage[pagenumbers]{cvpr} %

\usepackage[table,dvipsnames]{xcolor}

\def\eg{\emph{e.g}\onedot} 
\def\ie{\emph{i.e}\onedot} 
 
\def\etc{\emph{etc}\onedot} \def\vs{\emph{vs}\onedot}
\def\wrt{w.r.t\onedot}

\newcommand{\figref}[1]{Fig\onedot~\ref{#1}}

\newcommand{\tabref}[1]{Tab\onedot~\ref{#1}}

\definecolor{cvprblue}{rgb}{0.21,0.49,0.74}
\usepackage[pagebackref,breaklinks,colorlinks,citecolor=cvprblue]{hyperref}
\usepackage{multirow}
\usepackage{booktabs}
\usepackage{subcaption}
\usepackage{graphicx}
\usepackage{tabularx}
\usepackage{float}

\newlength\savewidth\newcommand\shline{\noalign{\global\savewidth\arrayrulewidth
  \global\arrayrulewidth 1pt}\hline\noalign{\global\arrayrulewidth\savewidth}}
\newcommand{\tablestyle}[2]{\setlength{\tabcolsep}{#1}\renewcommand{\arraystretch}{#2}\centering\footnotesize}

\def\eg{\emph{e.g}\onedot} 
\def\ie{\emph{i.e}\onedot} 
 
\def\etc{\emph{etc}\onedot} \def\vs{\emph{vs}\onedot}
\def\wrt{w.r.t\onedot}

\definecolor{darkpastelgreen}{rgb}{0.01, 0.75, 0.24}

\definecolor{amber}{rgb}{0.9, 0.65, 0.0}

\newcommand{\kmaxname}{kMaX-DeepLab\xspace}
\newcommand{\boxtomaskname}{box-kMaX\xspace}

\definecolor{baselinecolor}{gray}{.9}

\def\paperID{1518} %
\def\confName{CVPR}
\def\confYear{2024}

\begin{document}
\title{COCONut: Modernizing COCO Segmentation}

\author{Xueqing Deng
\;\;\;\;
Qihang Yu
\;\;\;\;
Peng Wang
\;\;\;\;
Xiaohui Shen
\;\;\;\;
Liang-Chieh Chen \\
ByteDance\\
\url{https://xdeng7.github.io/coconut.github.io/}  
}

\twocolumn[{%
\maketitle\centering
\vspace{-20pt}
\includegraphics[width=\linewidth]{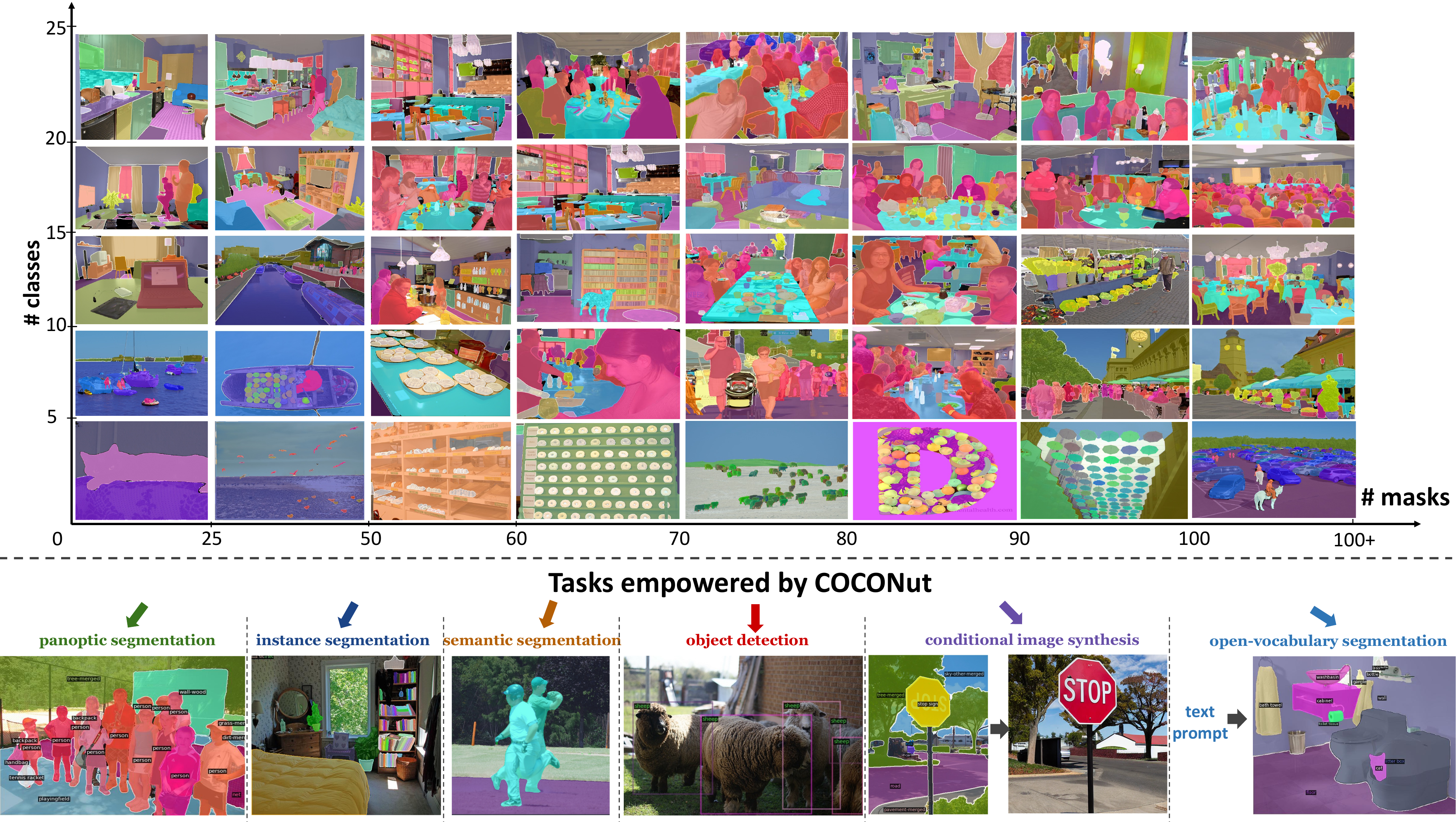}
    \vspace{-18pt}
  \captionof{figure}{
  \textbf{Overview of COCONut, the \textit{COCO} \textit{N}ext \textit{U}niversal segmen\textit{T}ation dataset:}
  \textit{Top:} COCONut, comprising images from COCO and Objects365, constitutes a diverse collection annotated with high-quality masks and semantic classes.
  \textit{Bottom:} COCONut empowers a multitude of image understanding tasks.
    }
    \vspace{10pt}
    \label{fig:teaser}
}]

\begin{abstract}
In recent decades, the vision community has witnessed remarkable progress in visual recognition, partially owing to advancements in dataset benchmarks.
Notably, the established COCO benchmark has propelled the development of modern detection and segmentation systems.
However, the COCO segmentation benchmark has seen comparatively slow improvement over the last decade.
Originally equipped with coarse polygon annotations for `thing' instances, it gradually incorporated coarse superpixel annotations for `stuff' regions, which were subsequently heuristically amalgamated to yield panoptic segmentation annotations.
These annotations, executed by different groups of raters, have resulted not only in coarse segmentation masks but also in inconsistencies between segmentation types.
In this study, we undertake a comprehensive reevaluation of the COCO segmentation annotations.
By enhancing the annotation quality and expanding the dataset to encompass 383K images with more than 5.18M panoptic masks, we introduce COCONut, the \textbf{COCO} \textbf{N}ext \textbf{U}niversal segmen\textbf{T}ation dataset.
COCONut harmonizes segmentation annotations across semantic, instance, and panoptic segmentation with meticulously crafted high-quality masks, and establishes a robust benchmark for all segmentation tasks.
To our knowledge, COCONut stands as the inaugural large-scale universal segmentation dataset, verified by human raters.
We anticipate that the release of COCONut will significantly contribute to the community's ability to assess the progress of novel neural networks.
\end{abstract}
    
\setcounter{table}{0}

\begin{figure*}[t!]
    \centering
    \includegraphics[width=\linewidth]{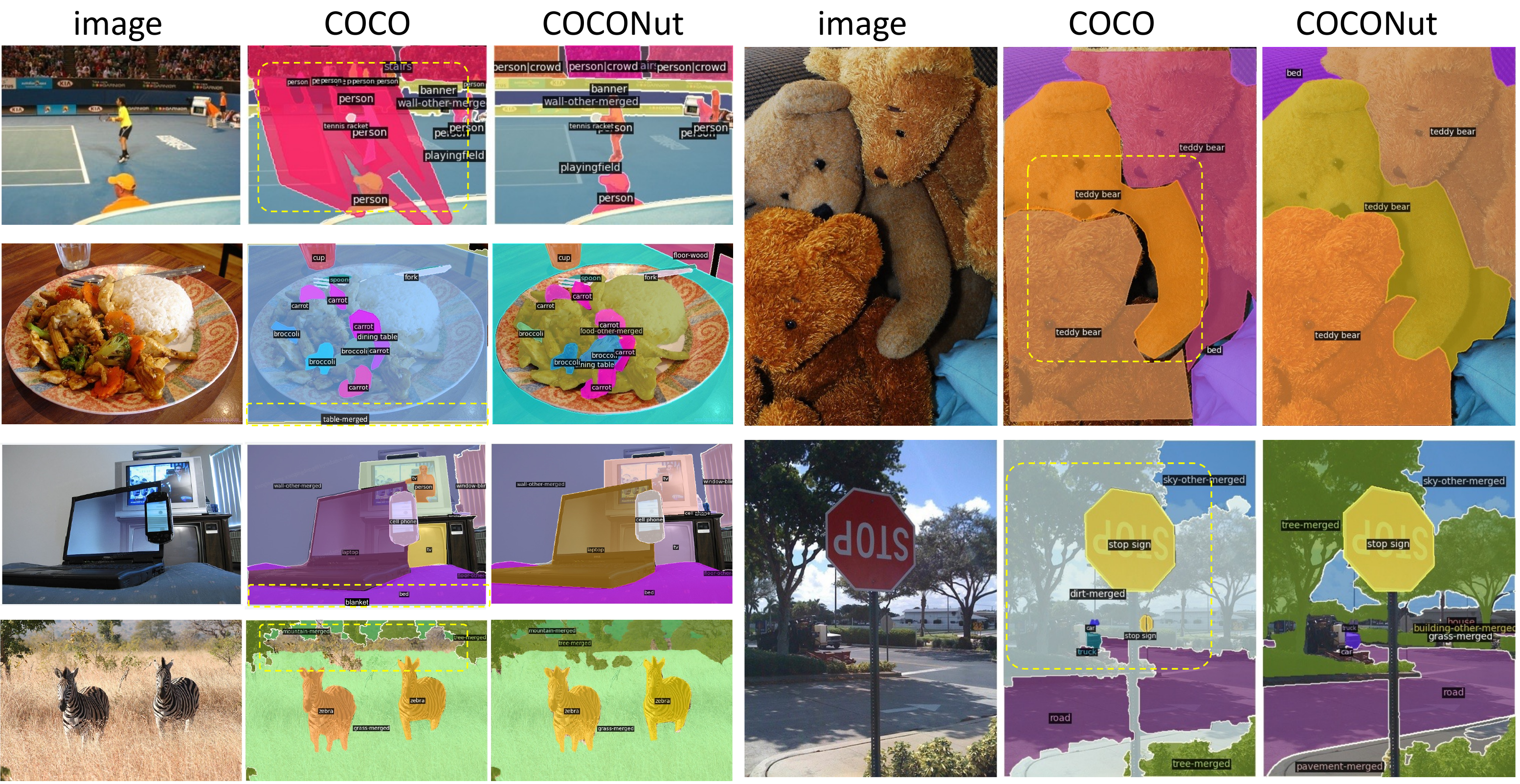}
    \vspace{-15pt}
    \caption{
    \textbf
    {Annotation Comparison:}
    We delineate erroneous annotations from COCO using \textcolor{amber}{yellow} dotted line boxes, juxtaposed with our COCONut corrected annotations.
    Common COCO annotation errors include over-annotations (\eg, `person crowd' erroneously extends into `playingfield'), incomplete mask fragments (\eg, `table-merged' and `blanket' are annotated in small isolated segments), missing annotations (\eg, `tree-merged' remains unannotated), coarse segmentations (especially noticeable in `stuff' regions annotated by superpixels and in `thing' regions by loose polygons), and wrong semantic categories (\eg, `tree-merged' is incorrectly tagged as `dirt-merged').
    }
    \vspace{5pt}
    \label{fig:vis_coco_coconut}
    
\end{figure*}
\section{Introduction}
\label{sec:intro}

Over the past decades, significant advancements in computer vision have been achieved, partially attributed to  the establishment of comprehensive benchmark datasets.
The COCO dataset~\cite{lin2014microsoft}, in particular, has played a pivotal role in the development of modern vision models, addressing a wide range of tasks such as object detection~\cite{hariharan2014simultaneous,girshick2015fast,ren2015faster,lin2017feature,carion2020end,qiao2021detectors,zhang2023dino}, segmentation~\cite{long2015fully,chendeeplabv1,chen2017deeplab,chen2018encoder,kirillov2019panopticfpn,cheng2020panoptic,wang2020axial,wang2021max,yu2022cmt,he2023maxtron,sun2024remax,shin2024video,yang2024polymax}, keypoint detection~\cite{he2017mask,guler2018densepose,papandreou2018personlab,sun2019deep}, and image captioning~\cite{chen2015microsoft,radford2021learning,yu2022coca}.
Despite the advent of large-scale neural network models~\cite{dosovitskiy2021vit,liu2022convnet,chen2024vitamin} and extensive datasets~\cite{shao2019objects365,kirillov2023sa-1b},
COCO continues to be a primary benchmark across various tasks, including image-to-text~\cite{yu2022coca,li2023blip2,liu2023visual} and text-to-image~\cite{saharia2022photorealistic,yu2022scaling} multi-modal models.
It has also been instrumental in the development of novel models, such as those fine-tuning on COCO for image captioning~\cite{rombach2022high,yu2022scaling} or open-vocabulary recognition~\cite{ghiasi2022scaling,kuo2022f,xu2023open,gu2023dataseg,yu2023convolutions,yu2023towards}.
However, nearly a decade since its introduction, the suitability of COCO as a benchmark for contemporary models warrants reconsideration. This is particularly pertinent given the potential nuances and biases embedded within the dataset, reflective of the early stages of computer vision research.

COCO's early design inevitably encompassed certain annotation biases, including issues like imprecise object boundaries and incorrect class labels (\figref{fig:vis_coco_coconut}).
While these limitations were initially acceptable in the nascent stages of computer vision research (\eg, bounding boxes are invariant to the coarsely annotated masks as long as the extreme points are the same~\cite{papadopoulos2017extreme}), the rapid evolution of model architectures has led to a performance plateau on the COCO benchmark\footnote{\url{https://paperswithcode.com/dataset/coco}}. This stagnation suggests a potential overfitting to the dataset's specific characteristics, raising concerns about the models' applicability to real-world data.
Furthermore, despite COCO's diverse annotations supporting various tasks, its annotation is neither exhaustive nor consistent.
This in-exhaustiveness is evident in the segmentation annotations, where instances of incomplete labeling are commonplace.
Additionally, discrepancies between semantic, instance, and panoptic annotations within the dataset present challenges in developing a comprehensive segmentation model.
Moreover, in the context of the ongoing shift towards even larger-scale datasets~\cite{schuhmann2022laion,gadre2023datacomp}, COCO's repository of approximately 120K images and 1.3M masks appears increasingly inadequate. This limitation hampers its utility in training and evaluating models designed to process and learn from substantially larger and more varied datasets.

\begin{table*}[t!]
\resizebox{\linewidth}{!}{%
\begin{tabular}{l|ccccccccccc}
 & COCONut & COCO-17~\cite{lin2014microsoft} & EntitySeg~\cite{qilu2023entity_dataset} & ADE20K~\cite{zhou2017scene} & Sama-COCO~\cite{samacoco} & 
LVIS~\cite{gupta2019lvis} &
Open Images~\cite{kuznetsova2020open} &
COCO-Stuff~\cite{caesar2018coco} & 
PAS-21~\cite{everingham2010pascal} &
PC-59~\cite{mottaghi2014role}\\
\shline
\# images (train/val/test) & 358K / 25K / - & 118K / 5K / 41K & 10K / 1.5K / -$\dagger$ & 20K / 2K / 3K$\ddagger$ & 118K / 5K / - & 100K / 20K / 40K &
944K / 13K / 40K &
118K / 5K / 41K & 1.4K / 1.4K / 1.4K & 5K / 5K / - \\
\# masks / image  &  13.2 / 17.4 / - & 11.2 / 11.3 / - & 16.8 / 16.4 / - & 13.4 / 15.1 / - & 9.0 / 9.5 / - & 12.7 / 12.4 / - & 2.8 / 1.8 / 1.8  & 8.6 / 8.9 / - & 2.5 / 2.5/ - & 4.9 / 4.8 / -  \\
\# masks  & 4.75M / 437K / -  & 1.3M / 57K / - & 0.17M / 24K / - & 0.27M / 30K / - & 1.07M / 47K / - & 1.27M / 0.24M / - & 2.7M / 25K / 74K & 1.02M / 44K / - & 3.6K / 3.6K / -  & 24K / 24K / - \\
\# thing classes  & 80 & 80 & 535 & 115 & 80 & 1203 & 350 & - & - & - \\
\# stuff classes & 53 & 53 & 109 & 35 & - & - & - & 91 & 21 & 59\\
\hline
panoptic segmentation & \checkmark & \checkmark & \checkmark & $\checkmark$ &  &  & &   &  &  \\
instance segmentation & \checkmark & \checkmark & \checkmark & $\triangle$ & \checkmark & \checkmark & \checkmark  & & \\
semantic segmentation & \checkmark & $\triangle$ & \checkmark & \checkmark & & & & \checkmark & \checkmark & \checkmark \\
object detection & \checkmark & \checkmark & $\triangle$ & $\triangle$ & \checkmark & \checkmark & \checkmark & & &\\
\end{tabular}
}
\vspace{-5pt}
\caption{
\textbf{Dataset Comparison:}
We compare existing segmentation datasets that focus on daily images (street-view images are not our focus).
The definition of `thing' and `stuff' classes are different across datasets, where the `stuff' classes are not annotated with instance identities.
$\dagger$: EntitySeg dataset comprises 33K images, of which only 11K are equipped with panoptic annotations.
$\ddagger$: ADE20K test server only supports semantic segmentation, and its panoptic annotations are derived by merging separately annotated instance and semantic segmentation maps, introducing minor inconsistencies between segmentation types.
$\triangle$: task supported, but not typically used.
}
\label{tab:dataset_comp}
\vspace{5pt}
\end{table*}

To modernize COCO segmentation annotations, we propose the development of a novel, large-scale universal segmentation dataset, dubbed COCONut for the \textbf{COCO}\textbf{N}ext \textbf{U}niversal segmen\textbf{T}ation dataset.
Distinct in its approach to ensuring high-quality annotations, COCONut features human-verified mask labels for 383K images.
Unlike previous attempts at creating large-scale datasets, which often compromise on label accuracy for scale~\cite{wang2023as-1b,schuhmann2022laion}, our focus is on maintaining human verification as a standard for dataset quality.
To realize this ambition, our initial step involves meticulously developing an assisted-manual annotation pipeline tailored for high-quality labeling on the subset of 118K COCO images, designated as COCONut-S split.
The pipeline benefits from modern neural networks (bounding box detector~\cite{ouyang2022nms} and mask segmenter~\cite{yu2022k,sun2023cfr}), allowing our annotation raters to efficiently edit and refine those proposals.
Subsequently, to expand the data size while preserving quality, we develop a data engine, leveraging the COCONut-S dataset as a high-quality training dataet to upgrade the neural networks.
The process iteratively generates various sizes of COCONut training sets, yielding COCONut-B (242K images and 2.78M masks), and COCONut-L (358K images and 4.75M masks). 

We adhere to a principle of consistency in annotation, aiming to establish a universal segmentation dataset (\ie, consistent annotations for all panoptic/instance/semantic segmentation tasks).
Additionally, COCONut includes a meticulously curated high-quality validation set, COCONut-val, comprising 5K images carefully re-labeled from the COCO validation set, along with an additional 20K images from Objects365~\cite{shao2019objects365} (thus, totally 25K images and 437K masks).

To summarize, our contributions are threefold:
\begin{itemize}
    \item We introduce COCONut, a modern, universal segmentation dataset that encompasses about 383K images and 5.18M human-verified segmentation masks. This dataset represents a significant expansion in both scale and quality of annotations compared to existing datasets. Additionally, COCONut-val, featuring meticulously curated high-quality annotations for validation, stands as a novel and challenging testbed for the research community.
    
    \item Our study includes an in-depth error analysis of the COCO dataset's annotations. This analysis not only reveals various inconsistencies and ambiguities in the existing labels but also informs our approach to refining label definitions. As a result, COCONut features ground-truth annotations with enhanced consistency and reduced label map ambiguity.

    \item With the COCONut dataset as our foundation, we embark on a comprehensive analysis. Our experimental results not only underscore the efficacy of scaling up datasets with high-quality annotations for both training and validation sets, but also highlight the superior value of human annotations compared to pseudo-labels.
    
\end{itemize}

\section{Related Work}
\label{sec:formatting}

In this work, we focus on segmentation datasets, featuring daily images (\tabref{tab:dataset_comp}).
A prime example of this is the COCO dataset~\cite{lin2014microsoft}, which has been a cornerstone in computer vision for over a decade. Initially, COCO primarily focused on detection and captioning tasks~\cite{chen2015microsoft}. Subsequent efforts have expanded its scope, refining annotations to support a wider array of tasks. For instance, COCO-Stuff~\cite{caesar2018coco} added semantic masks for 91 `stuff' categories, later integrated with instance masks to facilitate panoptic segmentation~\cite{kirillov2019panoptic}.
In addition to these expansions, several initiatives have aimed at enhancing the quality of COCO's annotations.
The LVIS dataset~\cite{gupta2019lvis} extends the number of object categories from 80 to 1,203, providing more comprehensive annotations for each image.
Similarly, Sama-COCO~\cite{samacoco} addresses the issue of low-quality masks in COCO by re-annotating instances at a finer granularity.
Beyond the COCO-related datasets, there are other notable datasets contributing to diverse research scenarios, including ADE20K~\cite{zhou2017scene}, PASCAL \cite{everingham2010pascal}, and PASCAL-Context~\cite{mottaghi2014role}. While these datasets have significantly advanced computer vision research, they still fall short in either annotation quality or quantity when it comes to meeting the demands for high-quality large-scale datasets.

In the realm of recent dataset innovations, SA-1B~\cite{kirillov2023sa-1b} stands out with its unprecedented scale, comprising 11M images and 1B masks.
However, a critical aspect to consider is the feasibility of human annotation at such an immense scale. Consequently, a vast majority (99.1\%) of SA-1B's annotations are machine-generated and lack specific class designations.
Additionally, its human annotations are not publicly released.
Contrasting with scaling dataset size, the EntitySeg dataset~\cite{qilu2023entity_dataset} prioritizes enhancing annotation quality. This dataset features high-resolution images accompanied by meticulously curated high-quality mask annotations. However, the emphasis on the quality of annotations incurs significant resource demands, which in turn limits the dataset's scope. As a result, EntitySeg encompasses a relatively modest collection of 33K images, of which only approximately one-third are annotated with panoptic classes.
Along the same direction of scaling up datasets, we present COCONut, a new large scale dataset with high quality mask annotations and semantic tags.

\section{Constructing the COCONut Dataset}
\label{sec:coconut_dataset}

In this section, we first revisit COCO's class map definition (Sec.~\ref{sec:class_def}) and outline our image sources and varied training data sizes (Sec.~\ref{sec:data_split}). 
The construction of COCONut centers on two key objectives: high quality and large scale. To achieve these, we establish an efficient annotation pipeline ensuring both mask quality and accurate semantic tags (Sec.~\ref{sec:anno_pipeline}). This pipeline facilitates scalable dataset expansion while upholding annotation quality (Sec.~\ref{sec:data_engine}).

\subsection{COCO's Class Map Definition}
\label{sec:class_def}

In alignment with the COCO panoptic set~\cite{kirillov2019panoptic}, COCONut  encompasses 133 semantic classes, with 80 categorized as `thing' and 53 as `stuff.'
Adopting the same COCO class map ensures backward compatibility, enabling the initial use of models trained on COCO-related datasets~\cite{lin2014microsoft,caesar2018coco,kirillov2019panoptic,samacoco} to generate pseudo labels in our annotation pipeline.

\begin{figure*}[ht!]
    \centering
    \includegraphics[width=\linewidth]{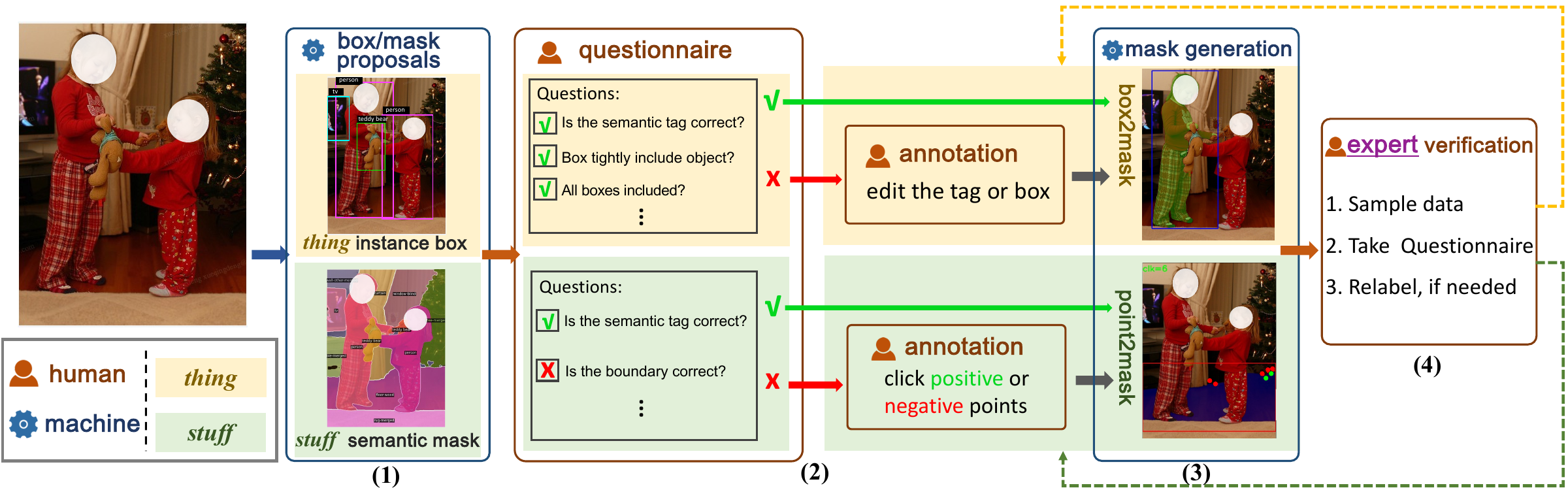}
    \vspace{-12pt}
    \caption{
    \textbf{Overview of the Proposed Assisted-Manual Annotation Pipeline:}
    To streamline the labor-intensive labeling task, our annotation pipeline encompasses four pivotal stages: (1) machine-generated pseudo labels, (2) human inspection and editing, (3) mask generation or refinement, and (4) quality verification. Acknowledging the inherent distinctions between `thing' and `stuff' classes, we systematically address these intricacies at each stage.
    Stage 1: Machines are employed to generate box and mask proposals for `thing' and `stuff', respectively.
    Stage 2: Raters assess the proposal qualities using a meticulously crafted questionnaire.
    For proposals falling short of requirements, raters can update them by editing boxes or adding positive/negative points for `thing' and `stuff', respectively.
    Stage 3: We utilize Box2Mask and Point2Mask modules to generate masks based on the inputs from stage 2.
    Stage 4: Experts perform a comprehensive verification of annotation quality, with relabeling done if the quality falls below our stringent standards.
    }
    \vspace{8pt}
    \label{fig:pipeline}
\end{figure*}

Notably, COCONut refines class map definitions compared to COCO, offering greater clarity in our annotation instruction protocol.
Building upon COCO's class map, we introduce additional definitions and instructions for labeling segmentation masks.
To mitigate the annotation confusion, we meticulously define label map details and provide clear instructions to our annotation raters.
For comprehensive definitions and annotation instructions for all 133 classes, please refer to the supplementary materials.

\subsection{Image Sources and Data Splits}
\label{sec:data_split}

The images comprising COCONut are sourced from public datasets.
Primarily, we aggregate images from the original COCO training and validation sets as well as its unlabeled set.
Additionally, we select approximately 136K images from Objects365 dataset~\cite{shao2019objects365}, each annotated with bounding boxes and containing at least one COCO class.
This comprehensive collection results in a total of 358K and 25K images for training  and validation, respectively. 
As illustrated in~\tabref{tab:dataset_definitions}, we meticulously define diverse training datasets for COCONut, spanning from 118K images to 358K images. 
COCONut-S (small) encompasses the same images as the original COCO training set, totaling 118K images.
We adopt COCO panoptic~\cite{lin2014microsoft} and Sama-COCO~\cite{samacoco} masks (high-quality instance segmentation annotations\footnote{\url{https://www.sama.com/sama-coco-dataset}}) as our starting point.
COCONut-B (base) incorporates additional images from the COCO unlabeled set, totaling 242K images.
Finally, with extra 116K images from the Objects365 dataset, COCONut-L (large) comprises 358K images.
Additionally, COCONut-val contains 5K images from the COCO validation set along with an additional 20K Objects365 images. 

\begin{table}[t!]
\scalebox{0.8}{
\tablestyle{2pt}{1.1}
\begin{tabular}{l|l|ccc}
dataset splits & \multicolumn{1}{c|}{image sources} & \#images & \#masks & \#masks/image \\
\shline
COCONut-S & COCO training set~\cite{lin2014microsoft} & 118K & 1.54M & 13.1 \\

COCONut-B & + COCO unlabeled set~\cite{lin2014microsoft} & 242K & 2.78M & 11.5 \\
COCONut-L & + subset of Objects365~\cite{shao2019objects365} & 358K & 4.75M & 13.2 \\ 
\hline
relabeled COCO-\textit{val} & COCO validation set~\cite{lin2014microsoft} & 5K & 67K & 13.4 \\
COCONut-\textit{val} & + subset of Objects365~\cite{shao2019objects365} & 25K & 437K & 17.4 \\ 
\end{tabular}
}
\vspace{-2pt}
\caption{
\textbf{Definition of COCONut Dataset Splits:}
Statistics are shown accumulatively. Notably, our COCONut-val contains large \#masks/image, preseting a more challenging testbed.
}
\label{tab:dataset_definitions}

\end{table}

\subsection{Assisted-Manual Annotation Pipeline}
\label{sec:anno_pipeline}

\quad\textbf{Annotation Challenges:}
The task of densely annotating images with segmentation masks, coupled with their semantic tags (\ie, classes), is exceptionally labor-intensive.
Our preliminary studies reveal that, on average, it takes one expert rater approximately 5 minutes to annotate a single mask.
Extrapolating this to annotate images at a scale of 10M masks would necessitate 95 years with just one expert rater.
Even with a budget to employ 100 expert raters, the annotation process would still require about a year to complete.
Given the extensive time and cost involved, this challenge underscores the need to explore a more effective and efficient annotation pipeline.

\textbf{Annotation Pipeline:}
In response to the challenges, we introduce the assisted-manual annotation pipeline, utilizing neural networks to augment human annotators.
As illustrated in~\figref{fig:pipeline}, the pipeline encompasses four key stages: (1) machine-generated prediction, (2) human inspection and editing, (3) mask generation or refinement, and (4) quality verification.
Recognizing the inherent differences between `thing' (countable objects) and `stuff' (amorphous regions), we meticulously address them at every stage.

\textbf{Machine-Generated Prediction:}
In handling `thing' classes, we utilize the bounding box object detector DETA~\cite{ouyang2022nms}, and for `stuff' classes, we deploy the mask segmenter kMaX-DeepLab~\cite{yu2022k}. This stage yields a set of box proposals for `thing' and mask proposals for `stuff'.

\textbf{Human Inspection and Editing:}
With the provided box and mask proposals, raters meticulously evaluate them based on a prepared questionnaire (\eg, Is the box/mask sufficiently accurate? Is the tag correct? Any missing boxes?)
The raters adhere to stringent standards during inspection to ensure proposal quality.
In cases where proposals fall short, raters are directed to perform further editing.
Specifically, for `thing' classes, raters have the flexibility to add or remove boxes along with their corresponding tags (i.e., classes).
In the case of `stuff' classes, raters can refine masks by clicking positive or negative points, indicating whether the points belong to the target instance or not.

\begin{figure}[t!]
    \centering
    \includegraphics[width=\linewidth]{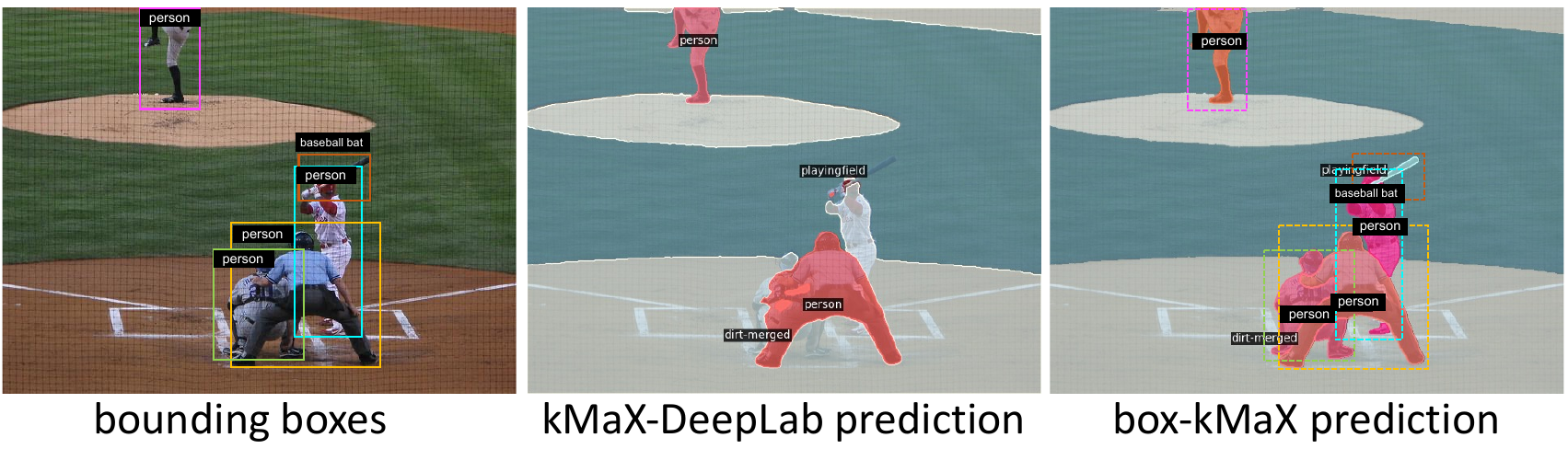}
        \vspace{-12pt}
    \caption{
    \textbf{Mask Prediction Comparison:}
    In contrast to kMaX-DeepLab, \boxtomaskname (Box2Mask module) leverages box queries, initialized with features pooled from the backbone within the box regions, enabling more accurate segmentation of  `thing' objects. Notably, kMaX-DeepLab falls short in capturing the challenging `baseball bat' and the heavily occluded `person' in the figure.
    }
    \label{fig:vis_boxkmax}
\end{figure}

\textbf{Mask Generation or Refinement:}
Utilizing the provided boxes and masks from the preceding stage, we employ the \textbf{Box2Mask} and \textbf{Point2Mask} modules to generate segmentation masks for `thing' and `stuff' classes, respectively.
The \textbf{Box2Mask} module extends kMaX-DeepLab, resulting in the \boxtomaskname model, which generates masks based on provided bounding boxes. 
This model incorporates additional box queries in conjunction with the original object queries.
The added box queries function similarly to the original object queries, except that they are initialized using features pooled from the backbone within the box regions (original object queries are randomly initialized).
As shown in~\figref{fig:vis_boxkmax}, leveraging object-aware box queries enables \boxtomaskname to effectively segment  `thing' objects with the provided bounding boxes.
The \textbf{Point2Mask} module utilizes the interactive segmenter CFR~\cite{sun2023cfr}, taking positive/negative points as input and optionally any initial mask (from either kMaX-DeepLab or the previous round's output mask). 
This stage allows us to amass a collection of masks generated from boxes and refined by points.

It is worth noting that there are other interactive segmenters that are also capable of generating masks using box and point as inputs (\eg, SAM~\cite{kirillov2023sa-1b}, SAM-HQ\cite{sam_hq}).
However, our analyses (in Sec.~\ref{sec:annotation_and_dataset}) indicate that the tools we have developed suffice for our raters to produce high-quality annotations.
The primary focus of our work is to conduct a comprehensive analysis between the original COCO dataset and our newly annotated COCONut. Improving interactive segmenters lies outside the scope of this study.

\begin{figure*}[ht!]
    \centering
    \includegraphics[width=\linewidth]{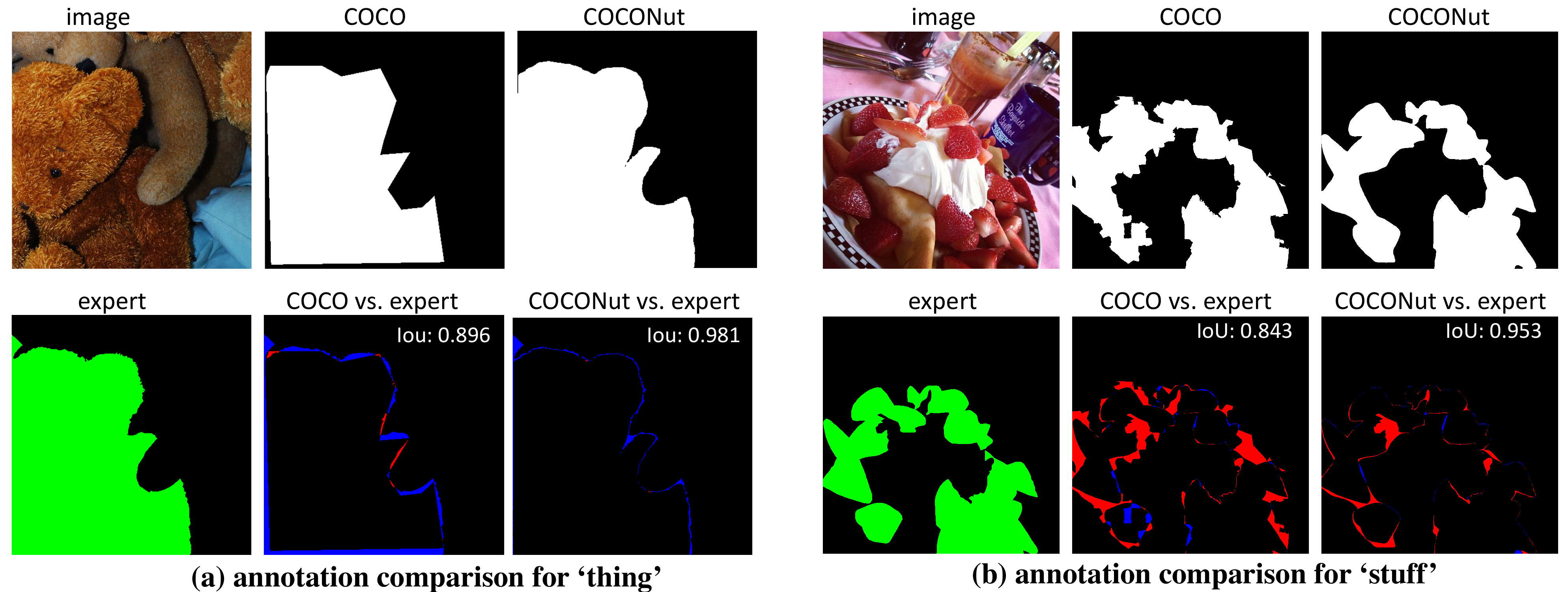}
    \vspace{-12pt}
     \caption{
    \textbf{Annotation Comparison:}
    We show annotations obtained by COCO, COCONut (Box2Mask for `thing' in (a) or Point2Mask for `stuff' in (b)), and our expert rater.
    COCONut's annotation exhibits sharper boundaries, closely resembling expert results, as evident from higher IoU values.
    The \textcolor{blue}{blue} and \textcolor{red}{red} regions correspond to extra and missing regions, respectively, compared to the expert \textcolor{darkpastelgreen}{mask}.
    }
    \label{fig:anno_tool_comp}
    \vspace{5pt}
\end{figure*}

\textbf{Quality Verification by Experts:}
Armed with the amassed masks from the preceding stage, we task \textit{expert raters} with quality verification.
Unlike the general human raters in stage 2, our expert raters boast extensive experience in dense pixel labeling (5 years of proficiency in Photoshop).
To manage the extensive volume of annotated masks with only two experts, we opt for a random sampling of 50\%.
The experts meticulously assess these masks, along with their associated tags, using the same carefully crafted questionnaire as in the previous stage.
Furthermore, recognizing the Box2Mask module's reliance on provided bounding boxes, we additionally instruct experts to verify the accuracy of box proposals, selecting 30\% samples for a thorough quality check.
Should any fall short of our stringent requirements, they undergo relabeling using the time-intensive Photoshop tool to ensure high annotation quality.

\subsection{Data Engine for Scaling Up Dataset Size}
\label{sec:data_engine}

\quad\textbf{Overview:}
With the streamlined assisted-manual annotation pipeline in place, we build a data engine to facilitate the dataset expansion.
Our data engine capitalizes on the annotation pipeline to accumulate extensive, high-quality annotations, subsequently enhancing the training of new neural networks for improved pseudo-label generation.
This positive feedback loop is iteratively applied multiple times.

\textbf{Data Engine:}
Machines play a crucial role in generating box/mask proposals (stage 1) and refined masks (stage 3) in the assisted-manual annotation pipeline. 
Initially, publicly available pre-trained neural networks are employed to produce proposals.
Specifically, DETA~\cite{ouyang2022nms} (utilizing a Swin-L backbone~\cite{liu2021swin} trained with Objects365~\cite{shao2019objects365} and COCO detection set~\cite{lin2014microsoft}) and kMaX-DeepLab~\cite{yu2022k} (featuring a ConvNeXt-L backbone~\cite{liu2022convnet} trained with COCO panoptic set~\cite{kirillov2019panoptic}) are utilized to generate box and mask proposals for `thing' and `stuff', respectively.
The Point2Mask module (built upon CFR~\cite{sun2023cfr}) remains fixed throughout the COCONut construction, while the Box2Mask module (\boxtomaskname, a variant of kMaX-DeepLab using box queries) is trained on COCO panoptic set.
The annotation pipeline initially produces the COCONut-S dataset split. 
Subsequently, COCONut-S is used to re-train kMaX-DeepLab and box-kMaX, enhancing mask proposals for ‘stuff’ and Box2Mask capabilities, respectively.
Notably, DETA and the Point2Mask module are not re-trained, as DETA is already pre-trained on a substantial dataset, and CFR exhibits robust generalizability.
The upgraded neural networks yield improved proposals and mask generations, enhancing the assisted-manual annotation pipeline and leading to the creation of COCONut-B.
This process is iterated to generate the final COCONut-L, which also benefits from the ground-truth boxes provided by Objects365.

\section{Annotation and Data Engine Analysis}
\label{sec:annotation_and_dataset}

In this section, we scrutinize the annotations produced through our proposed assisted-manual annotation pipeline (Sec.~\ref{sec:annotation_analysis}). Subsequently, we delve into the analysis of the improvement brought by our data engine (Sec.~\ref{sec:data_engine_analysis}).

\subsection{Annotation Analysis}
\label{sec:annotation_analysis}

\begin{table}[t!]
\centering
\subfloat[
\textbf{Annotation Agreement}
\label{tab:iou_agreement}
]{
\centering
\begin{minipage}{0.5\linewidth}{\begin{center}
\tablestyle{2pt}{1.1}
\scalebox{0.95}{
\begin{tabular}{l|cc}
 & `thing' & `stuff'\\
\shline
expert-1 \vs expert-2 & 98.1\% & 97.3\% \\
\hline
raters \vs experts & 96.3\% & 96.7\% \\
\end{tabular}
}
\end{center}}\end{minipage}
}
\subfloat[
\textbf{Annotation Speed}
\label{tab:speed_comp}
]{
\centering
\begin{minipage}{0.5\linewidth}{\begin{center}
\tablestyle{2pt}{1.1}
\scalebox{0.95}{
\begin{tabular}{l|cc}
 & `thing' & `stuff' \\
 \shline
 purely-manual & 10 min & 5 min \\
 \hline
 assisted-manual  & 10 sec & 42 sec \\
\end{tabular}
}
\end{center}}\end{minipage}
}
\vspace{-5pt}
\caption{
\textbf{Annotation Analysis:}
(a) Our two experts and raters demonstrate a high level of agreement in their annotations.
(b) The assisted-manual pipeline expedites the annotation.
}
\label{tab:annotation_analysis}
\end{table}

\begin{figure*}[t!]
    \centering
    \includegraphics[width=\linewidth]{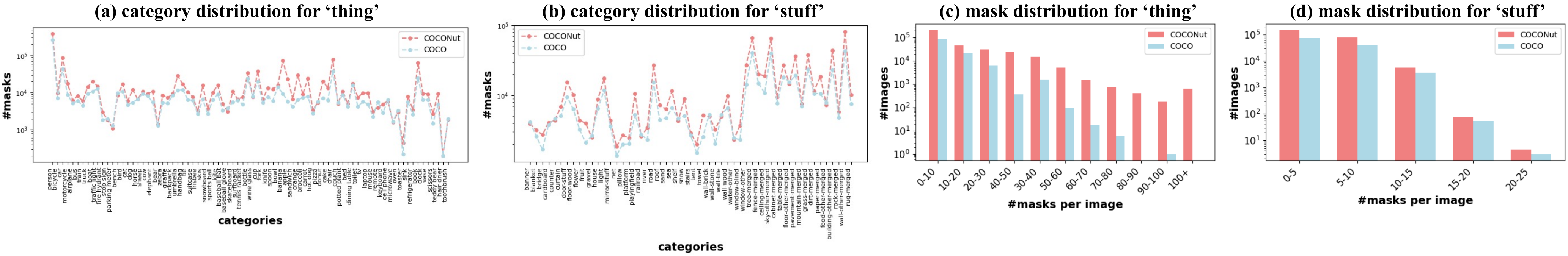}
    \vspace{-15pt}
    \caption{
    \textbf{Dataset Statistics:}
    In subfigures (a) and (b), depicting category distributions for `thing' and `stuff', COCONut consistently displays a higher number of masks across all categories compared to COCO.
    Subfigures (c) and (d) show mask distribution for `thing' and `stuff', respectively, demonstrating that COCONut contains a greater number of images with a higher density of masks per image.
    }
    \label{fig:dataset_stats}
\end{figure*}

\begin{table}[t!]
\centering
\subfloat[
\textbf{Non-Pass Rate in Stage 2}
\label{tab:non_pass_rate}
]{
\centering
\begin{minipage}{0.5\linewidth}{\begin{center}
\tablestyle{2pt}{1.1}
\scalebox{0.95}{
\begin{tabular}{l|cc}

constructed dataset       & mean & median \\
\shline
COCONut-S    & 78\%  & 75\%   \\
COCONut-B    & 51\%  & 55\%  \\
COCONut-L    & 43\%  & 45\% \\
\end{tabular}

}
\end{center}}\end{minipage}
}
\subfloat[
\textbf{\#Rounds of Relabeling in Stage 4}
\label{tab:relabeling_round}
]{
\centering
\begin{minipage}{0.5\linewidth}{\begin{center}
\tablestyle{2pt}{1.1}
\scalebox{0.95}{
\begin{tabular}{l|cc}
constructed dataset      & mean & median  \\
 \shline
COCONut-S    & 2.4 & 2    \\
COCONut-B    &  0.8 & 1  \\
COCONut-L    & 0.5  & 1 \\
\end{tabular}
\label{tab:round_relabel}
}
\end{center}}\end{minipage}
}
\vspace{-12pt}
\caption{
\textbf{Data Engine Analysis:} During the creation of the current dataset split, the mask proposals stem from models trained on datasets from preceding stages, such as COCONut-S utilizing proposal models from COCO, and so forth.
}
\label{tab:data_engin_analysis}
\end{table}

\quad\textbf{Assisted-Manual \vs Purely-Manual:} 
We conduct a thorough comparison in this study between annotations generated by our assisted-manual and purely-manual annotation pipelines.
Our assessment is based on two metrics: annotation quality and processing speed.
     
The purely-manual annotation pipeline involves two in-house experts, each with over 5 years of experience using Photoshop for labeling dense segmentation maps.
They received detailed instructions based on our annotation guidelines and subsequently served as tutorial training mentors for our annotation raters.
Additionally, they played a crucial role in the quality verification of masks during stage 4.

To conduct the ``agreement'' experiments, we randomly selected 1000 segmentation masks and tasked our two in-house experts with annotating each mask. An ``agreement'' was achieved when both annotations exhibited an IoU (Intersection-over-Union) greater than 95\%.
As presented in~\tabref{tab:iou_agreement}, our experts consistently demonstrated a high level of agreement in annotating both `thing' and `stuff' masks.
Comparatively, minor disparities were observed in the annotations provided by our raters, highlighting their proficiency.
Additionally, Tab.~\ref{tab:speed_comp} showcases the annotation speed. 
The assisted-manual pipeline notably accelerates the annotation process by editing boxes and points, particularly beneficial for `thing' annotations.
Annotating `stuff', however, involves additional time due to revising the coarse superpixel annotations by COCO.
Finally, Fig.~\ref{fig:anno_tool_comp} presents annotation examples from COCO, our experts, and COCONut (our raters with the assisted-manual pipeline), underscoring the high-quality masks produced.

\subsection{Data Engine Analysis}
\label{sec:data_engine_analysis}

\begin{figure}[t!]
    \centering
    \includegraphics[width=\linewidth]{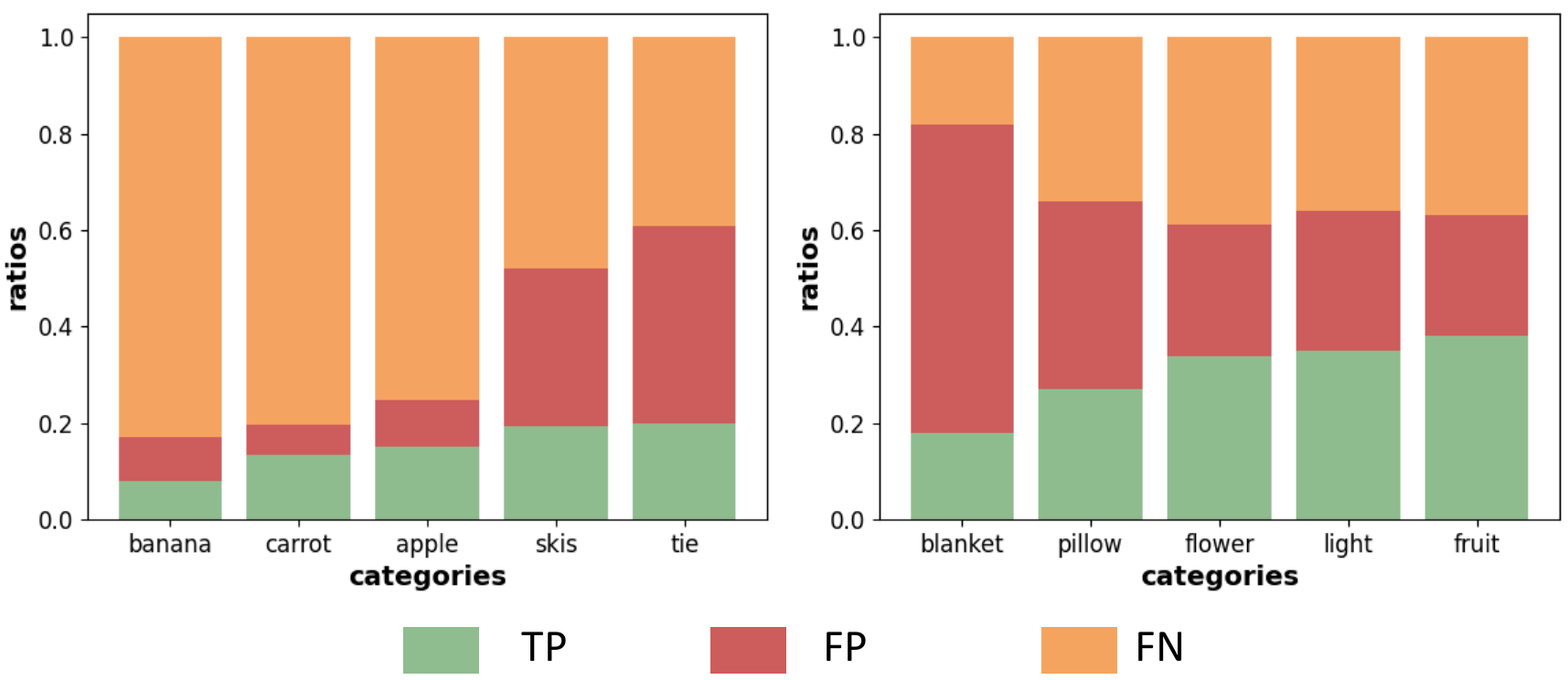}
    \vspace{-12pt}
    \caption{\textbf{Top 5 Disagreed Categories Between COCO-val and relabeled COCO-val:}  
    COCO-val is treated as the prediction, while relabeled COCO-val serves as ground truth.
    The comparison showcases True Positive (TP), False Positive (FP), and False Negative (FN) rates for both 'thing' (left) and 'stuff' (right).
    } 
    \label{fig:val_set_comp}
\end{figure}

The data engine enhances neural networks using annotated high-quality data, resulting in improved pseudo masks and decreased workload for human raters.
To measure its impact, we present non-pass rates in stage 2 human inspection. These rates indicate the percentage of machine-generated proposals that failed our questionnaire's standards and required further editing.
\tabref{tab:non_pass_rate} demonstrates that including more high-quality training data improves non-pass rates, signifying enhanced proposal quality.
Furthermore, \tabref{tab:relabeling_round} showcases the number of relabeling rounds in stage 4 expert verification, reflecting additional iterations required for annotations failing expert verification. Consistently, we observed reduced relabeling rounds with increased inclusion of high-quality training data.

\section{Dataset Statistics}
\label{sec:dataset_statistics}

\quad\textbf{Class and Mask Distribution:}
Fig.~\ref{fig:dataset_stats} depicts the category and mask distribution within COCONut.
Panels~(a) and~(b) demonstrate that COCONut surpasses COCO in the number of masks across all categories.
Additionally, panels~(c) and~(d) feature histograms depicting the frequency of `masks per image'. These histograms highlight a notable trend in COCONut, indicating a higher prevalence of images with denser mask annotations compared to COCO.

\begin{table}[t!]
\tablestyle{4pt}{1.1}
\begin{tabular}{c|ccc|ccc}
& PQ   & SQ   & RQ   & PQ\textsuperscript{bdry} & SQ\textsuperscript{bdry} & RQ\textsuperscript{bdr}     \\
\shline
all & 67.1 & 86.2 & 77.4 & 59.2 & 79.4 & 74.5  \\
thing & 65.0 & 86.0 & 75.2 & 58.6 & 80.7 & 72.4  \\
stuff & 70.2 & 86.5 & 80.8 &  60.1 & 77.3 & 77.6 \\
\end{tabular}
\vspace{-5pt}
\caption{
\textbf{Quantitative Comparison Between COCO-val and relabeled COCO-val:} COCO-val serves as the prediction, contrasting with relabeled COCO-val as the ground-truth. 
}
\label{tab:comp_coco}
\end{table}

\begin{figure}[t!]
    \centering
    \includegraphics[width=0.9\linewidth]{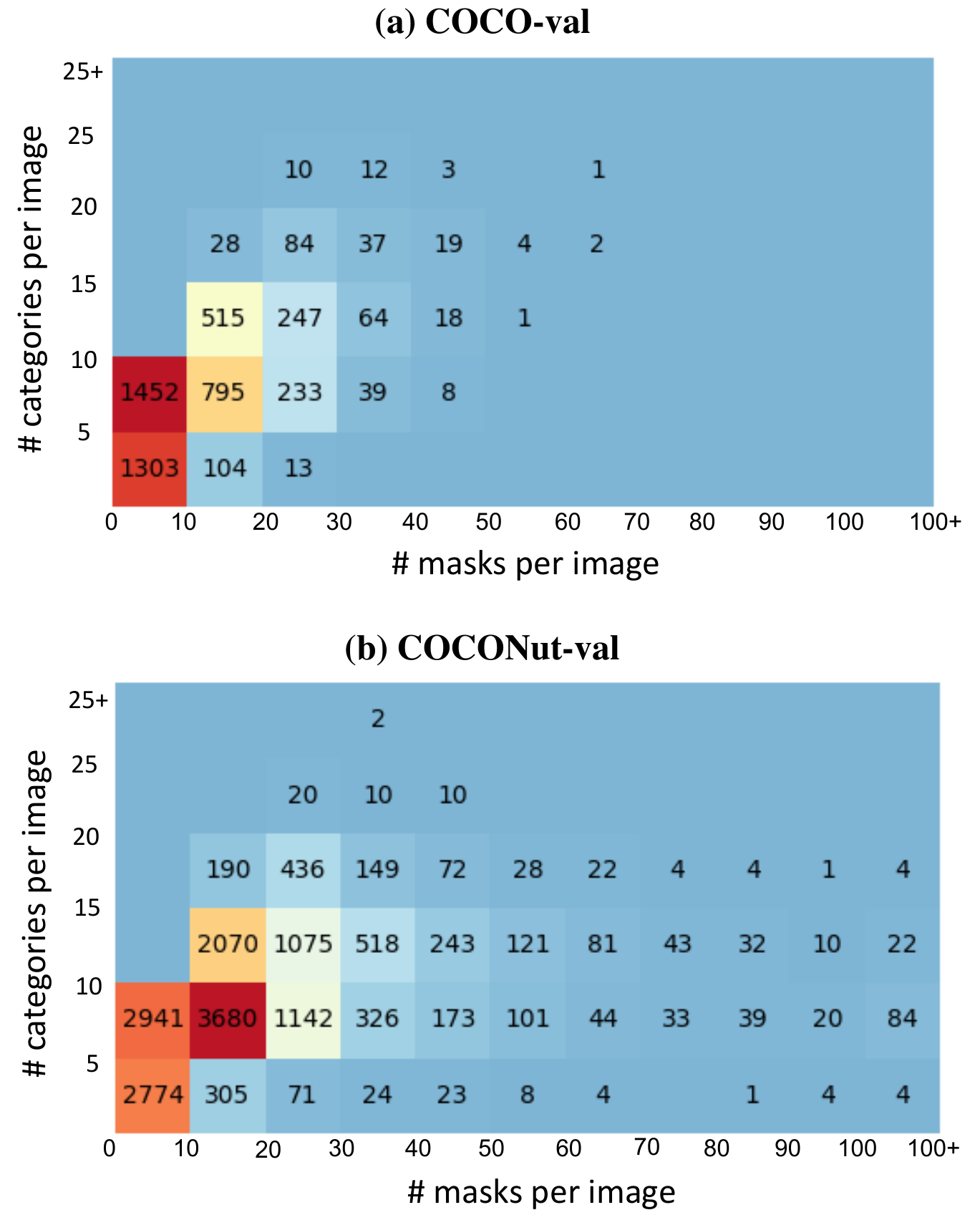}
        \vspace{-5pt}
    \caption{
    \textbf{Mask and Class Frequency Distribution:}
    COCONut-val introduces a more challenging testbed compared to the original COCO-val. It features a greater number of images that contain higher quantities of both masks and distinct categories per image.}

    \vspace{5pt}
    \label{fig:mask_freq_val_coco}
\end{figure}

\textbf{COCO-val \vs relabeled COCO-val:}
We conducted a comparative analysis between the original COCO-val annotations and our relabeled COCO-val.
Exploiting the Panoptic Quality (PQ) metric, we employed its True Positive (TP), False Positive (FP), and False Negative (FN) rates to assess each category.
TP signifies agreement between annotations, while FP and FN highlight additional or missing masks, respectively.
In Fig.~\ref{fig:val_set_comp}, we present the top 5 categories displaying discrepancies for both `thing' and `stuff'.
All these categories exhibit notably low TP rates, indicating substantial differences between COCO-val and our relabeled version.
In `thing' categories, high FN rates (around 0.8) are observed for `banana', `carrot', and `apple', suggesting numerous missing masks. Conversely, `stuff' categories exhibit high FP rates for `blanket' and `pillow', indicating numerous small isolated masks, echoing our earlier findings regarding `bed' and `blanket' conflicts, as depicted in Fig.~\ref{fig:vis_coco_coconut} (row 3).
Finally, \tabref{tab:comp_coco} provides a quantitative analysis comparing COCO-val and our relabeled COCO-val.
The results emphasize the notable divergence between the two sets, underscoring our dedicated efforts to improve the annotation quality of validation set.
The discrepancy is particularly evident in boundary metrics~\cite{cheng2021boundary}.
Notably, the divergence in stuff SQ\textsuperscript{bdry} reflects our enhancements to the original `stuff' annotations by superpixels~\cite{achanta2012slic,caesar2018coco}.

\textbf{COCONut-val (a new challenging testbed):}
To augment our relabeled COCO-val, we introduced an additional 20K annotated images from Objects365, forming COCONut-val.
Fig.~\ref{fig:mask_freq_val_coco} illustrates 2D histograms comparing COCO-val and COCONut-val, where we count the number of images \wrt their \#masks and \#categories per image.
The figure showcases that COCO-val annotations are concentrated around a smaller number of masks and categories, whereas COCONut-val demonstrates a broader distribution, with more images having over 30 masks. On average, COCONut-val boasts 17.4 masks per image, significantly exceeding COCO-val's average of 11.3 masks.

\begin{table}[t!]
\centering
\scalebox{0.78}{
\tablestyle{2pt}{1.1}
\begin{tabular}{c|c||ccc|ccc|ccc}
          &              & \multicolumn{3}{c|}{COCO-val} & \multicolumn{3}{c|}{relabeled COCO-val} & \multicolumn{3}{c}{COCONut-val} \\

 backbone & training set & PQ  & $\text{AP}^{\text{mask}}$ & mIoU & PQ  & $\text{AP}^{\text{mask}}$ & mIoU & PQ  & $\text{AP}^{\text{mask}}$ & mIoU \\
\shline
\multirow{4}{*}{ResNet50} & COCO & 53.3 & 39.6 & 61.7  & 55.1 & 40.6 & 63.9 & 53.1& 37.1 & 62.5 \\
 & COCONut-S & 51.7 & 37.5 & 59.4 & 58.9 & 44.4 & 64.4  & 56.7 &41.2 & 63.6\\
 & COCONut-B & 53.4 & 39.3 & 62.6 & 60.2 & 45.2 & 65.7  & 58.1 &42.9 &64.7  \\
 & COCONut-L & 54.1 & 40.2 & 63.1 & 60.7 & 45.8& 66.1  & 60.7 & 44.8 & 68.3 \\
\hline
\multirow{4}{*}{ConvNeXt-L} & COCO & 57.9 & 45.0 & 66.9 
 & 60.4 & 46.4 & 69.9 & 58.3 & 44.1 & 66.4 \\
 & COCONut-S & 55.9 & 41.9 & 66.1 & 64.4 & 50.8 & 71.4 &  59.4 &45.7 &67.8\\
 & COCONut-B &57.8 & 44.8 & 66.6 & 64.9 & 51.2 & 71.8 & 61.3 & 46.5 & 69.5 \\
 & COCONut-L & 58.1 & 45.3 & 67.3 & 65.1 &51.4 & 71.9  & 62.7 & 47.6 & 70.6\\

\end{tabular}
}
\vspace{-5pt}
\caption{
\textbf{Training Data and Backbones:}
The evaluations are conducted on three different validation sets: original COCO-val, relabeled COCO-val (by our raters), and COCONut-val. 
}
\label{tab:panseg}
\end{table}

\vspace{5pt}
\section{Discussion}
\label{sec:results}

In light of the COCONut dataset, we undertake a meticulous analysis to address the following inquiries.
We employ kMaX-DeepLab~\cite{yu2022k} throughout the experiments, benchmarked with several training and validation sets.

\begin{figure}[t!]
    \centering
    \includegraphics[width=\linewidth]{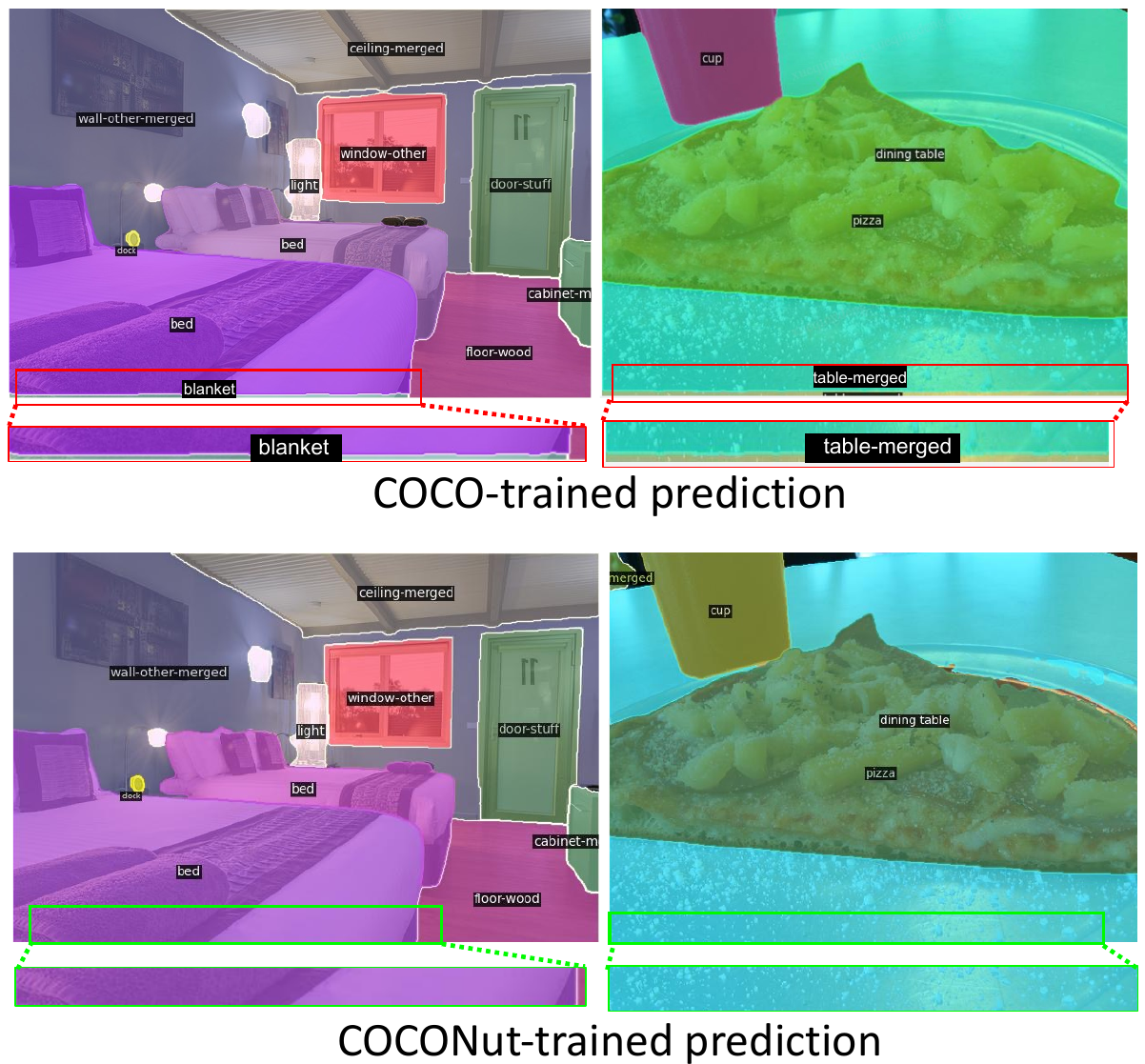}
    \vspace{-15pt}
    \caption{
    \textbf{Influence of Training Data on Predictions:}
    We present predictions from two models: one trained on original COCO (top) and the other on COCONut (bottom).
    \textit{Top}: The COCO-trained model predicts a small isolated mask, influenced by the biases inherent in the COCO coarse annotations (\eg, see Fig.~\ref{fig:vis_coco_coconut}, row 3). \textit{Bottom}: The COCONut-trained model does not predict small isolated masks, thanks to the  meticulously crafted annotations. 
    Best zoomed-in. 
    }
    \label{fig:pred_comp}
\end{figure}

\textbf{COCO encompasses only 133 semantic classes. Is an extensive collection of human annotations truly necessary?}
We approach this query from two vantage points: the training and validation sets.
Tab.~\ref{tab:panseg} showcases consistent improvements across various backbones (ResNet50~\cite{he2016deep} and ConvNeXt-L~\cite{liu2022convnet}) and three evaluated validation sets (measured in PQ, AP, and mIoU) as the training set size increases from COCONut-S to COCONut-L. 
Interestingly, relying solely on the original small-scale COCO training set yields unsatisfactory performance on both relabeled COCO-val and COCONut-val sets, emphasizing the need for more human annotations in training.
Despite annotation biases between COCO and COCONut (Fig.~\ref{fig:pred_comp}), training with COCONut-B achieves performance akin to the original COCO training set on the COCO validation set, hinting that a larger training corpus might mitigate inter-dataset biases. 

Shifting our focus from the training set to the validation set, the results in Tab.~\ref{tab:panseg} indicate performance saturation on both COCO-val and relabeled COCO-val as the training set expands from COCONut-B to COCONut-L.
This saturation phenomenon in COCO-val, consisting of only 5K images, is also observed in the literature\footnote{\url{https://paperswithcode.com/dataset/coco}}, suggesting its inadequacy in evaluating modern segmenters. 
Conversely, the newly introduced COCONut-val, comprising 25K images with denser mask annotations, significantly improves benchmarking for models trained with varied data amounts. This outcome underscores the significance of incorporating more human-annotated, challenging validation images for robust model assessment.
Therefore, the inclusion of additional human-annotated images is pivotal for both training and validation, significantly impacting the performance of modern segmentation models.

\begin{table}[t!]
\centering
\scalebox{0.78}{
\tablestyle{2pt}{1.1}
\begin{tabular}{c|c||ccc|ccc|ccc}
          &              & \multicolumn{3}{c|}{COCO-val} & \multicolumn{3}{c|}{relabeled COCO-val} & \multicolumn{3}{c}{COCONut-val} \\
 backbone & training set & PQ\textsuperscript  & $\text{AP}^{\text{mask}}$ & mIoU & PQ\textsuperscript  & $\text{AP}^{\text{mask}}$ & mIoU & PQ\textsuperscript  & $\text{AP}^{\text{mask}}$ & mIoU\\
\shline
\multirow{5}{*}{ConvNeXt-L} & COCO & 57.9 & 45.0 & 66.9 & 60.4 & 46.4 &  69.9 & 58.3 & 44.1 & 66.4 \\ 
 & COCO-B\textsubscript{M} & 58.0 & 44.9 & 67.1 & 60.7 & 46.3 & 70.5 & 58.5 & 44.2 & 66.4 \\
 \cline{2-11}
 & COCONut-S & 55.9 & 41.9 & 66.1 & 64.4 & 50.8 & 71.4 & 59.4 & 45.7 & 67.8 \\
 & COCONut-B\textsubscript{M} & 56.2 & 41.8 & 66.3 & 64.5 &  50.9 & 71.4 & 59.5 & 45.3 & 67.7 \\
 & COCONut-B & 57.8 & 44.8 & 66.6 & 64.9 & 51.2 & 71.8 &  61.3 &  46.5& 69.5\\
\end{tabular}
}
\caption{
\textbf{Pseudo-Labels \vs Human Labels:}
COCO-B\textsubscript{M} comprises the original COCO training set plus the machine pseudo-labeled COCO unlabeled set.
COCONut-B\textsubscript{M} contains COCONut-S and machine pseudo-labeled COCO unlabeled set (in contrast to the fully human-labeled COCONut-B).
}
\vspace{5pt}
\label{tab:pseudo}
\end{table}

\textbf{Are pseudo-labels a cost-effective alternative to human annotations?}
While expanding datasets using machine-generated pseudo-labels seems promising for scaling models trained on large-scale data, its effectiveness remains uncertain.
To address this, we conducted experiments outlined in~\tabref{tab:pseudo}.
Initially, leveraging a checkpoint (row 1: 57.9\% PQ on COCO-val), we generated pseudo-labels for the COCO unlabeled set, augmenting the original COCO training set to create the COCO-B\textsubscript{M} dataset.
Surprisingly, training on COCO-B\textsubscript{M} resulted in only a marginal 0.1\% PQ improvement on COCO-val, consistent across all tested validation sets (1st and 2nd rows in the table).

We hypothesized that the annotation quality of the pre-trained dataset might influence pseudo-label quality.
To investigate, we then utilized a different checkpoint (row 3: 64.4\% PQ on relabeled COCO-val) to generate new pseudo-labels for the COCO unlabeled set. 
Combining these with COCONut-S produced the COCONut-B\textsubscript{M} dataset, yet still yielded a mere 0.1\% PQ improvement on the relabeled COCO-val.
Notably, employing the fully human-labeled COCONut-B resulted in the most significant improvements (last row in the table).
Our findings suggest limited benefits from incorporating pseudo-labels. Training with pseudo-labels seems akin to distilling knowledge from a pre-trained network~\cite{hinton2015distilling}, offering minimal additional information for training new models.

\vspace{5pt}
\section{Visualization of COCONut Annotations}
\label{sec:coconut_vis}

\begin{figure*}[t!]
    \centering
    \includegraphics[width=\linewidth]{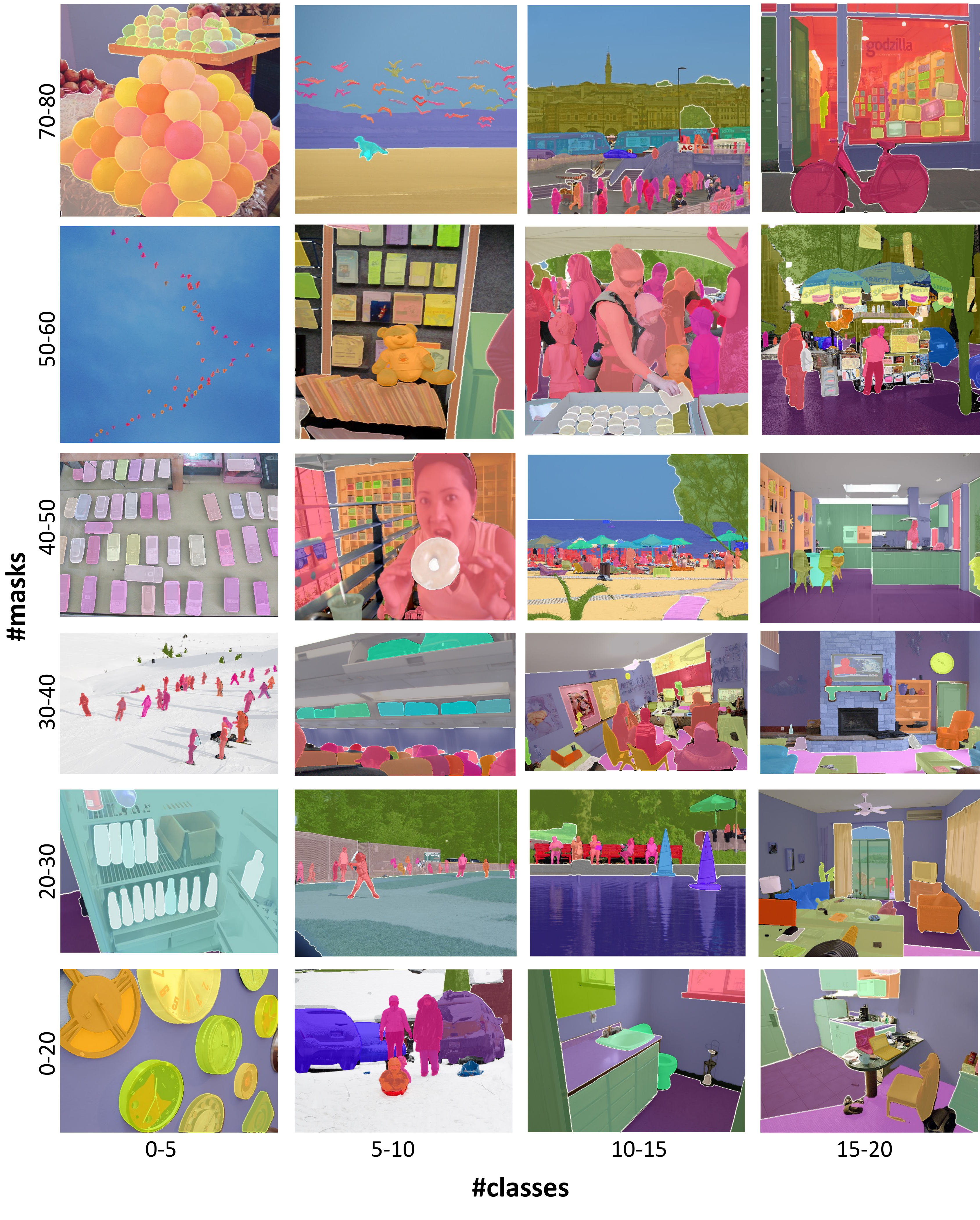}
    \caption{
    \textbf{Visualization of COCONut Annotations:}
    This figure demonstrates COCONut annotations with images sourced from COCO unlabeled set images. COCONut provides annotations comprising a large number of classes and masks.
    }
    \label{fig:vis_coconut}
\end{figure*}

\begin{figure*}[t!]
    \centering
    \includegraphics[width=\linewidth]{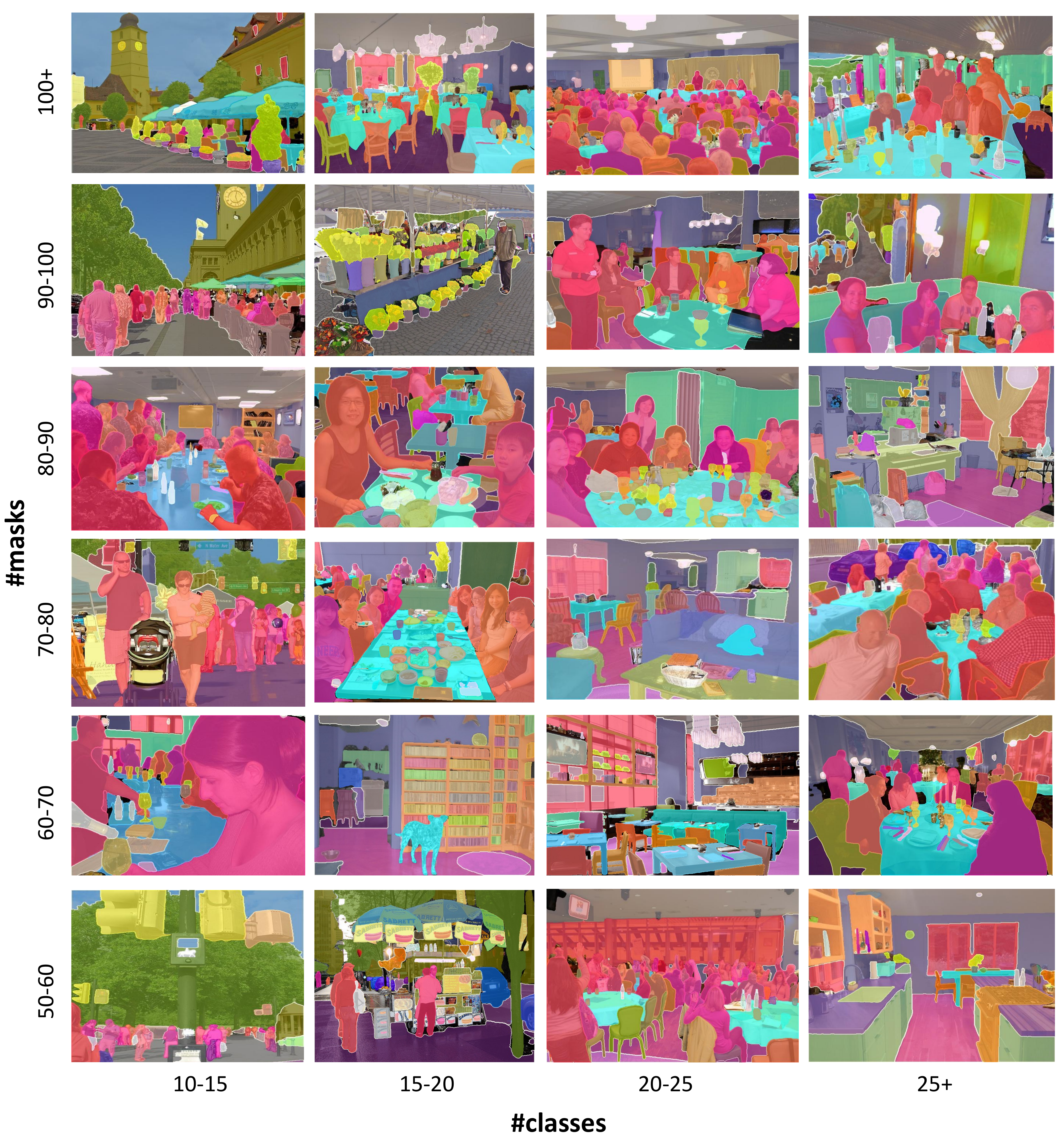}
    \caption{
    \textbf{Visualization of COCONut Annotations:}
    This figure showcases COCONut annotations using images sourced from both the COCO unlabeled set and Objects365 images.
    The inclusion of Objects365 images enriches COCONut annotations by introducing a wider variety of classes and masks compared to the COCO images.
    }
    \label{fig:vis_coconut_dense_mask}
\end{figure*}

\begin{figure*}[ht!]
    \centering
    \includegraphics[width=\linewidth]{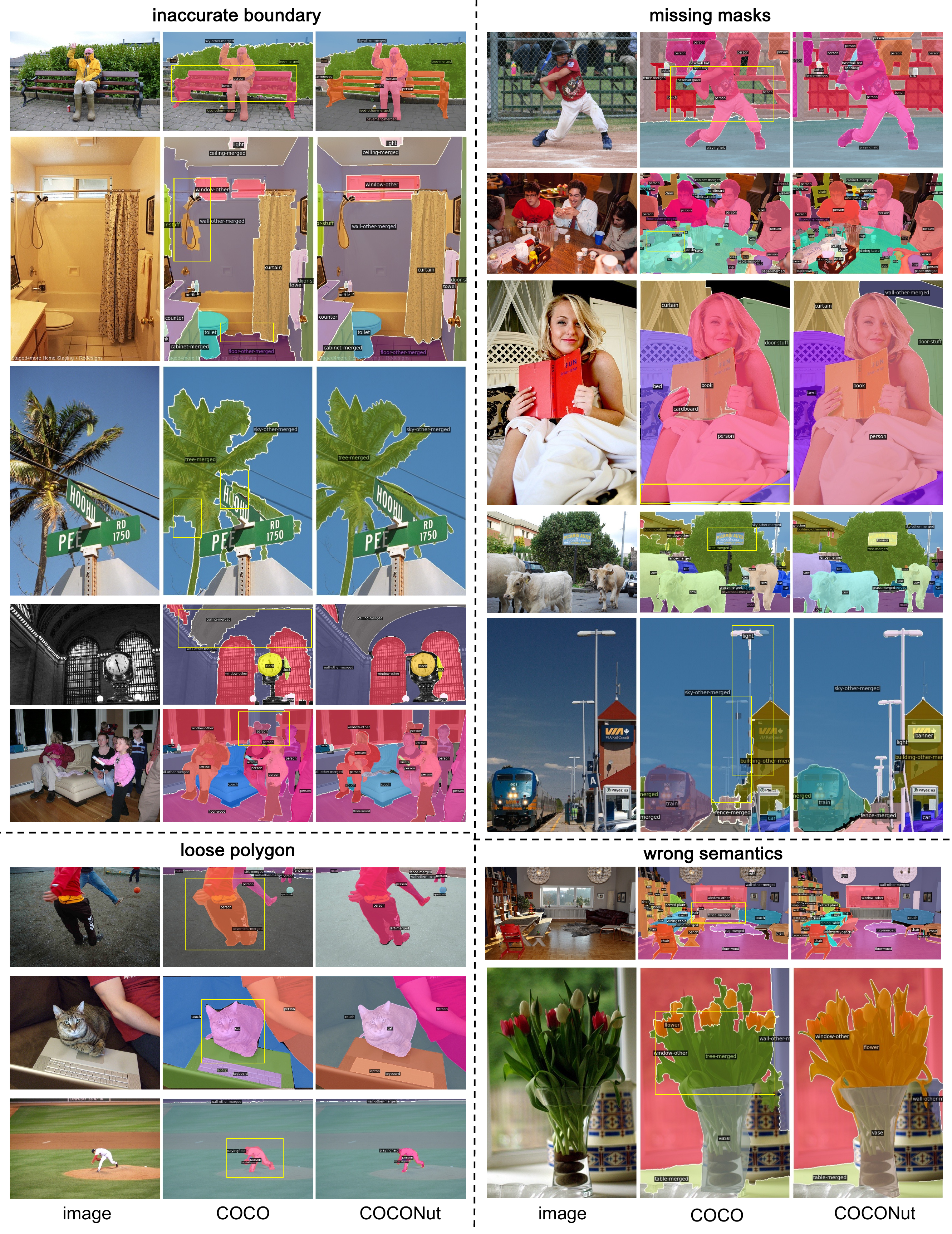}
    \vspace{-22pt}
    \caption{
    \textbf{Visualization Comparison Between COCO and COCONut:}
    COCONut effectively mitigates the annotations errors by COCO.
    The \textcolor{amber}{yellow} boxes highlight the erroneous areas in COCO.
    } 
    \label{fig:comp_coco_coconut}
\end{figure*}

We present annotation visualizations for COCONut dataset.
Specifically, Fig.~\ref{fig:vis_coconut} and Fig.~\ref{fig:vis_coconut_dense_mask} demonstrate the COCONut annotations for images sourced from COCO unlabeled set~\cite{lin2014microsoft} and Objects365~\cite{shao2019objects365}.
As shown in the figures, COCONut provides annotations comprising a large number of classes and masks.
Notably, the inclusion of Objects365 images enriches COCONut annotations by introducing a wider variety of classes and masks compared to the COCO images.
Finally, Fig.~\ref{fig:comp_coco_coconut} compares the COCO and COCONut annotations, where the common errors of COCO (\eg, inaccurate boundary, loose polygon, missing masks, and wrong semantics) are all corrected in COCONut annotations.

\clearpage

\appendix
\section*{Appendix}
\label{sec:appendix}
In the supplementary materials, we provide additional information, as listed below.

\begin{itemize}
    \item Sec.~\ref{sec:addition_figs} presents additional annotation visualizations.
    \item Sec.~\ref{sec:sup_results} provides more benchmarking results on COCONut.
    \item Sec.~\ref{sec:label_map} provides the details of our label map definition and annotation rules.
\end{itemize}

\section{Additional Annotation Visualizations}

\label{sec:addition_figs}
We present additional annotation visualizations for COCONut dataset.

\begin{figure*}
    \centering
    \includegraphics[width=\linewidth]{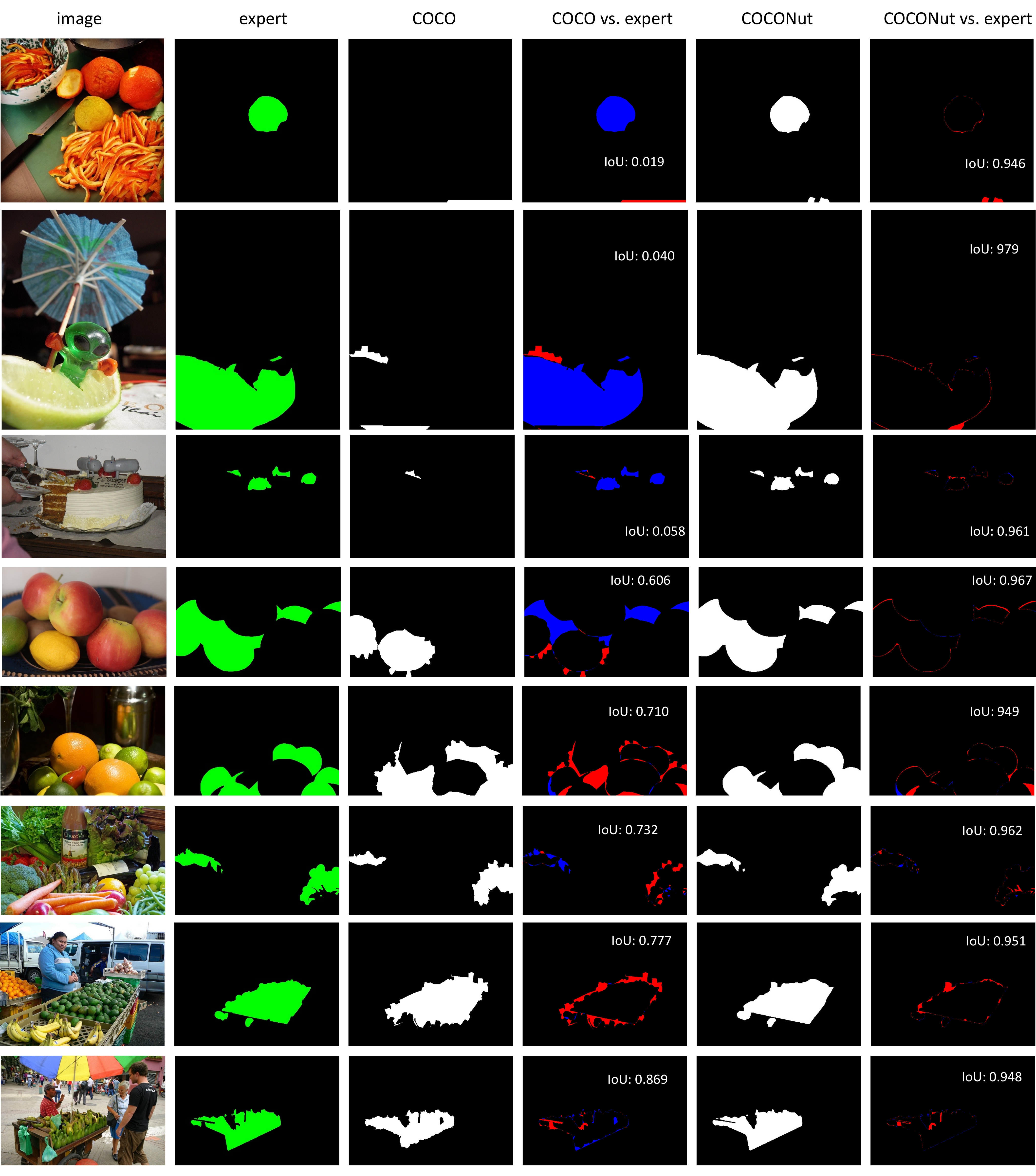}
    \vspace{-15pt}
    \caption{
    \textbf{Annotation Comparison:}
    We show annotations obtained by COCO, COCONut with Point2Mask for `stuff', and our expert rater.
    COCONut's annotation exhibits sharper boundaries, closely resembling expert results, as evident from higher IoU values.
    The \textcolor{blue}{blue} and \textcolor{red}{red} regions correspond to extra and missing regions, respectively, compared to the expert \textcolor{darkpastelgreen}{mask}.}
    \label{fig:comp_expert_stuff}
\end{figure*}

\begin{figure*}[ht!]
    \centering
    \includegraphics[width=\linewidth]{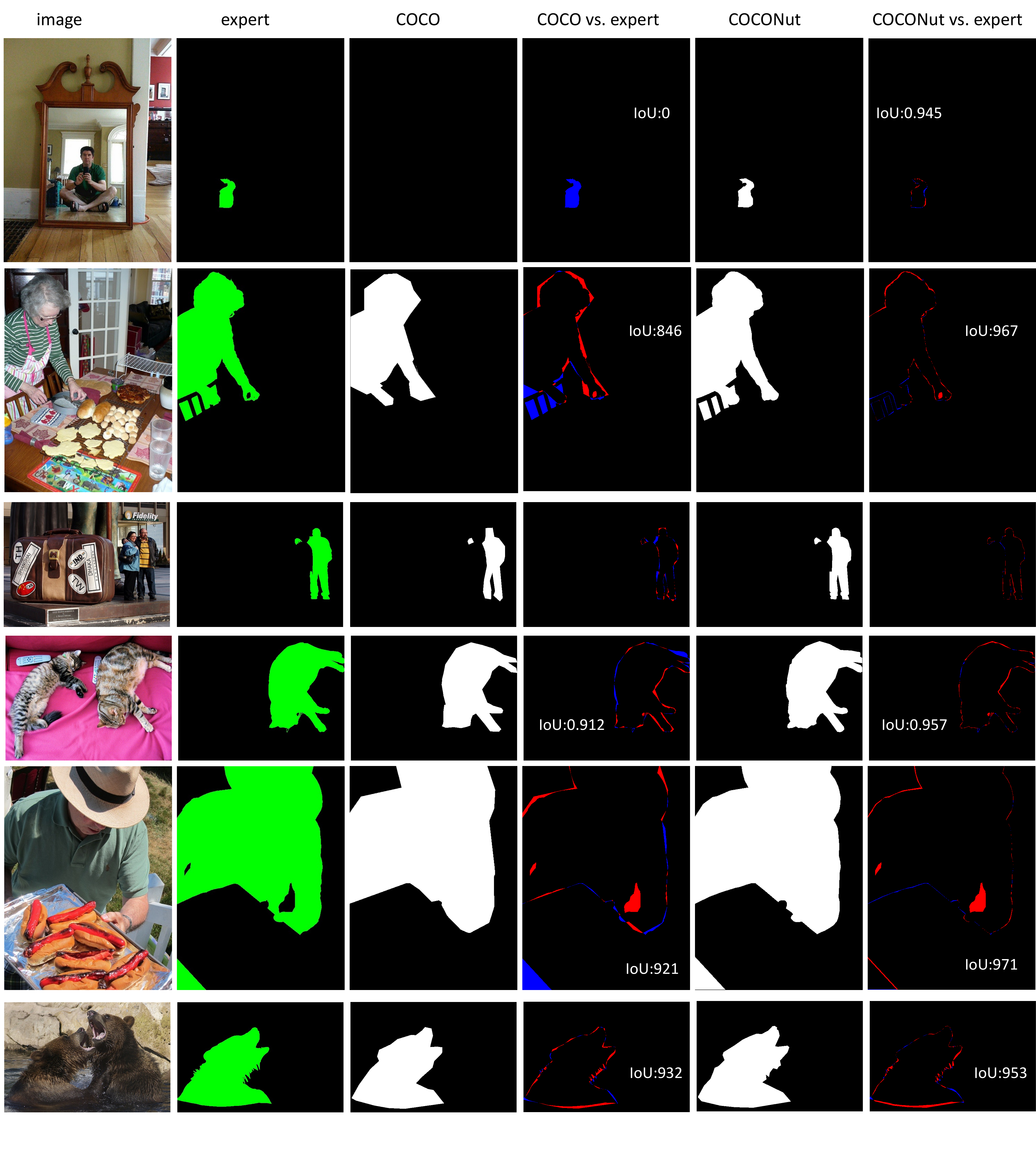}
    \caption{\textbf{Annotation Comparison:}
    We show annotations obtained by COCO, COCONut with Box2Mask for `thing', and our expert rater.
    COCONut's annotation exhibits sharper boundaries, closely resembling expert results, as evident from higher IoU values.
    The \textcolor{blue}{blue} and \textcolor{red}{red} regions correspond to extra and missing regions, respectively, compared to the expert \textcolor{darkpastelgreen}{mask}.} 
    \label{fig:comp_expert_thing}
\end{figure*}

\begin{figure*}[ht!]
    \centering
    \includegraphics[width=\linewidth]{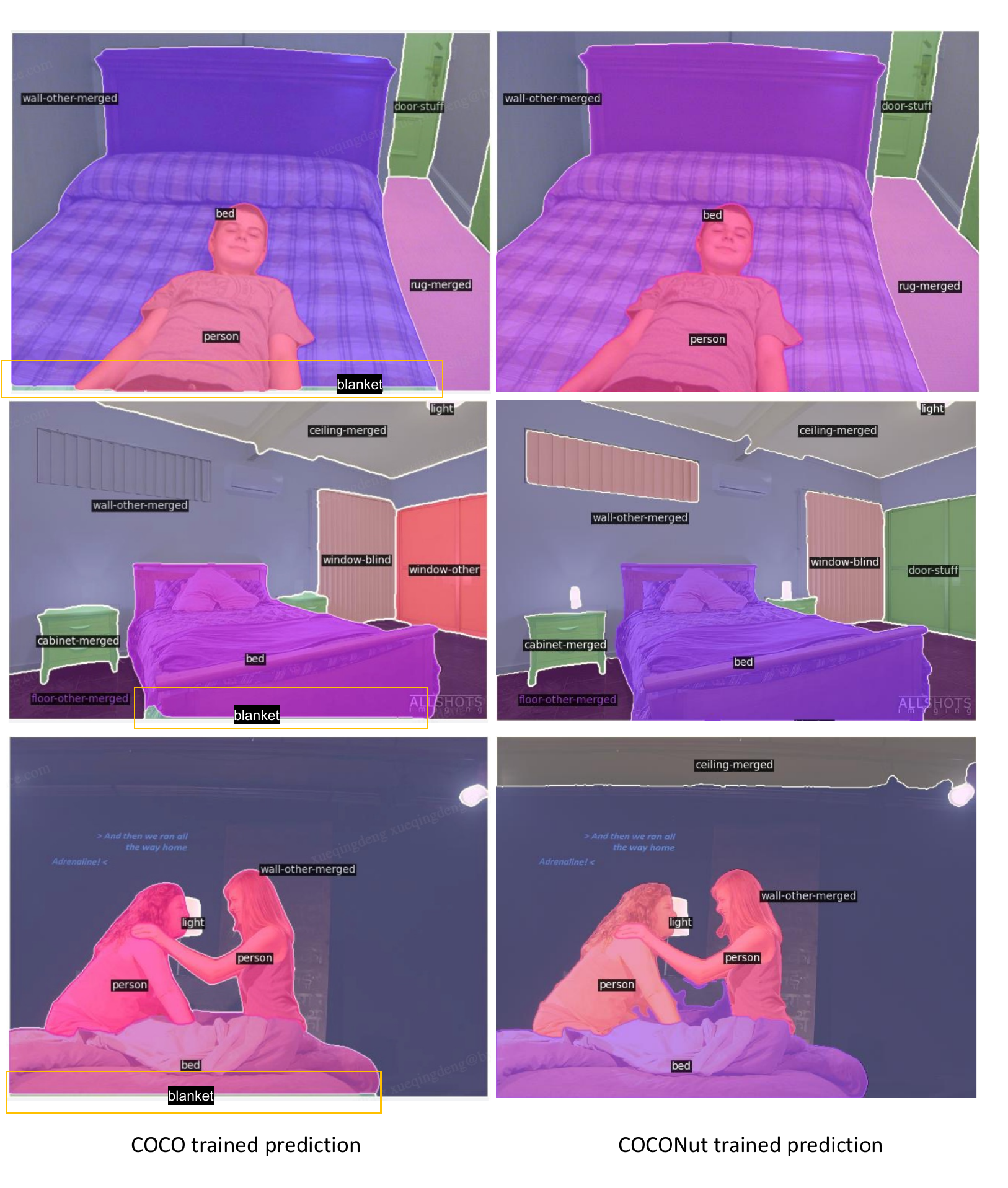}
    \caption{
    \textbf{Influence of Training Data on Predictions:} We present predictions from two models: one trained on original COCO (left) and the other on COCONut (right). The COCO-trained model predicts a small isolated mask, influenced by the biases inherent in the COCO coarse annotations.
    }
    \label{fig:prediction_bias}
\end{figure*}

\begin{figure*}[ht!]
    \centering
    \includegraphics[width=\linewidth]{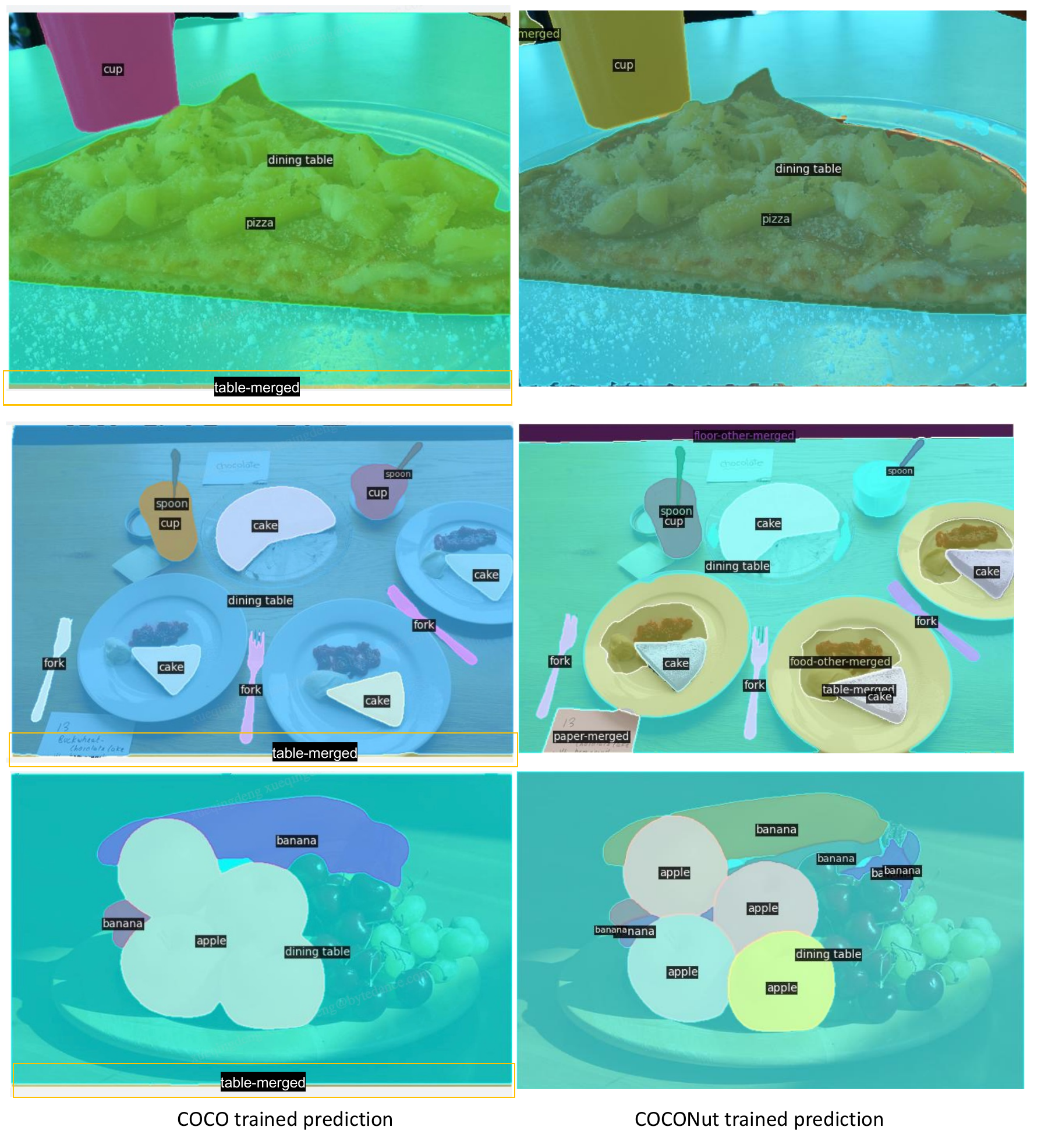}
    \caption{
    \textbf{Influence of Training Data on Predictions:} We present predictions from two models: one trained on original COCO (left) and the other on COCONut (right). The COCO-trained model predicts a small isolated mask, influenced by the biases inherent in the COCO coarse annotations.
    }
    \label{fig:prediction_bias_table}
\end{figure*}

In particular, Fig.~\ref{fig:comp_expert_stuff} and Fig.~\ref{fig:comp_expert_thing} provide more annotation comparisons between COCO, COCONut, and our expert raters.
Fig.~\ref{fig:prediction_bias} and Fig.~\ref{fig:prediction_bias_table} provide more visualizations of prediction bias introduced by training data.

\section{Additional Experimental Results}
\label{sec:sup_results}
\label{sec:appendeix_}
In this section, we outline the training and evaluation protocols utilized to benchmark COCONut across multiple tasks and the corresponding results in Sec.~\ref{sec:implementation_details} and Sec.~\ref{sec:various_tasks}, respectively.

\subsection{Training and Evaluation Protocols}
\label{sec:implementation_details}

\textbf{Training Protocol:}
COCONut undertakes benchmarking across various tasks, encompassing panoptic segmentation~\cite{kirillov2019panoptic}, instance segmentation~\cite{hariharan2014simultaneous}, semantic segmentation~\cite{He2004CVPR}, object detection~\cite{everingham2010pascal}, and open-vocabulary segmentation~\cite{ding2022open,ghiasi2022scaling}.
The \kmaxname~\cite{yu2022k,weber2021deeplab2}, tailored for universal segmentation, serves as the primary framework for panoptic, instance, and semantic segmentation in our experiments.
Object detection relies on the DETA framework~\cite{ouyang2022nms}, while open-vocabulary segmentation utilizes the FC-CLIP framework~\cite{yu2023convolutions}.

Throughout each task, we strictly adhere to the training hyper-parameters defined by the respective frameworks, utilizing ResNet50~\cite{he2016deep}, Swin-L~\cite{liu2021swin}, and ConvNeXt-L~\cite{liu2022convnet} as the backbones.

\textbf{Evaluation Protocol:}
When evaluating each task, we follow official settings meticulously.
For panoptic, instance, and semantic segmentation tasks, metrics such as panoptic quality (PQ)~\cite{kirillov2019panoptic}, AP\textsuperscript{mask}~\cite{lin2014microsoft}, and mean Intersection-over-Union (mIoU)~\cite{everingham2010pascal} are reported.
Bounding box detection performance is measured using the AP\textsuperscript{box} metric.
In line with prior methodologies~\cite{ding2022open,xu2023open}, open-vocabulary segmentation results undertake zero-shot evaluation on other segmentation datasets~\cite{everingham2010pascal,mottaghi2014role,zhou2017scene}.

\subsection{COCONut Empowers Various Tasks}
\label{sec:various_tasks}
In this section, we show the results for task-specific models trained on COCONut datasets including panoptic segmentation, instance segmentation, semantic segmentation,  object detection, semantic mask conditional image synthesis.

\begin{table}[t!]
\tablestyle{2pt}{1.1}
\scalebox{0.75}{
\begin{tabular}{c|c|c||c|c|c}
  &  &  & COCO-val & relabeled COCO-val & COCONut-val \\
 method  & backbone & training set & PQ & PQ & PQ \\
\shline

\multirow{8}{*}{\kmaxname} & \multirow{4}{*}{ResNet50}  & COCO  & 53.3 & 55.1 & 53.1 \\
 & & COCONut-S  & 51.7 & 58.9 & 56.7 \\
 & & COCONut-B  & 53.4 & 60.2 & 58.1 \\
 & & COCONut-L  & 54.1 & 60.7 & 60.7 \\
\cline{2-6}
& \multirow{4}{*}{ConvNeXt-L} & COCO      & 57.9 &  60.4 & 58.3  \\
 & & COCONut-S  & 55.9 & 64.4 & 59.4 \\
 & & COCONut-B  & 57.8 & 64.9 & 61.3 \\
 & & COCONut-L  & 58.1 & 65.1 & 62.7 \\
\end{tabular}
}
\caption{
\textbf{Benchmarking Task-Specific Panoptic Segmentation Models:}
kMaX-DeepLab is trained with \textit{panoptic} segmentation annotations across various training and validation sets.
}
\label{tab:benchmark_pan_seg}
\end{table}

\begin{table}[t!]
\tablestyle{2pt}{1.1}
\scalebox{0.75}{
\begin{tabular}{c|c|c||c|c|c}
  &  &  & COCO-val & relabeled COCO-val & COCONut-val \\
 method  & backbone & training set & AP\textsuperscript{mask} & AP\textsuperscript{mask} & AP\textsuperscript{mask} \\
\shline
\multirow{8}{*}{\kmaxname} & \multirow{4}{*}{ResNet50}  & COCO  & 44.1 & 44.6 & 41.9 \\
 & & COCONut-S  & 40.9 & 49.2 & 44.9 \\
 & & COCONut-B  & 41.2& 50.3 & 46.2 \\
 & & COCONut-L  & 41.4 & 50.9 & 47.1 \\
\cline{2-6}
& \multirow{4}{*}{ConvNeXt-L} & COCO      & 49.2 & 50.2 & 47.1  \\
 & & COCONut-S  & 45.5 & 55.8 & 51.2 \\
 & & COCONut-B  & 46.4 &  56.7 & 52.9 \\
 & & COCONut-L  & 47.0 & 57.0 & 53.8 \\
\end{tabular}
}
\caption{
\textbf{Benchmarking Task-Specific Instance Segmentation Models:}
kMaX-DeepLab is trained with
\textit{instance} segmentation annotations across various training and validation sets.
}
\label{tab:benchmark_inst_seg}
\end{table}

\begin{table}[t!]
\tablestyle{2pt}{1.1}
\scalebox{0.75}{
\begin{tabular}{c|c|c||c|c|c}
  &  &  & COCO-val & relabeled COCO-val & COCONut-val \\
 method  & backbone & training set & mIoU & mIoU & mIoU \\
\shline
\multirow{8}{*}{\kmaxname} & \multirow{4}{*}{ResNet50}  & COCO  & 59.5 & 64.6 & 62.9 \\
 & & COCONut-S  & 59.3 & 66.4 & 65.1  \\
 & & COCONut-B  & 63.5 & 67.3 & 66.5 \\
 & & COCONut-L  & 64.2 & 68.0 & 67.8 \\
\cline{2-6}
& \multirow{4}{*}{ConvNeXt-L} & COCO      & 67.1 & 70.9 & 68.1  \\
 & & COCONut-S  & 66.1 & 71.9 & 69.9 \\
 & & COCONut-B  & 67.4 &  72.4 & 71.3 \\
 & & COCONut-L  & 67.5 & 72.7 & 72.6 \\
\end{tabular}
}
\caption{
\textbf{Benchmarking Task-Specific Semantic Segmentation Models:}
kMaX-DeepLab is trained with
\textit{semantic} segmentation annotations across various training and validation sets.
}
\label{tab:benchmark_sem_seg}
\end{table}

\begin{table}[t!]
\centering
\tablestyle{1pt}{1}
\scalebox{0.8}{
\begin{tabular}{c|c||c|c|c}
  &     & \multicolumn{3}{c}{evaluation set (mIoU)} \\
backbone & training dataset & COCO-val & relabeled COCO-val &  COCONut-val  \\
\shline

\multirow{4}{*}{ViT-Adapter-B} & COCO & 61.2 & 64.5 & 61.8 \\
& COCONut-S & 60.6 & 66.0 & 64.9 \\
& COCONut-B & 61.3 & 66.9 &  66.3 \\
& COCONut-L & 62.4 & 67.7 & 67.1 \\
\hline
\multirow{4}{*}{ViT-Adapter-L} & COCO & 66.6 & 69.9 & 67.5 \\
& COCONut-S & 65.2 & 71.0 & 69.5 \\
& COCONut-B & 66.4 & 72.1 & 70.7 \\
& COCONut-L & 67.2  & 72.3 & 71.0 \\
\end{tabular}
}
\vspace{-8pt}
\caption{
\textbf{Benchmarking plain ViT backbone for Semantic Segmentation:}
Mask2Former w/ ViT-Adapter is trained with \textit{semantic} segmentation annotations.
}
\label{tab:rebuttal_vitadapter}
\end{table}

\textbf{Panoptic Segmentation:}
In~\tabref{tab:benchmark_pan_seg}, we benchmark \kmaxname   on the task of panoptic segmentation.
The results are the same as Tab.~6 in the main paper, where a panoptic model is evaluated on all three segmentation metrics.

\begin{table*}[ht!]
\centering
\tablestyle{6pt}{1.1}
\scalebox{1}{
\begin{tabular}{c|c|c||ccc||cccc}
       &          &               & \multicolumn{3}{c||}{ADE20K-150} & A-847 & PC-459 & PC-59 & PAS-21\\
method & backbone & training data & PQ & $\text{AP}^{\text{mask}}$ & mIoU & mIoU& mIoU & mIoU & mIoU\\
\shline
\multirow{4}{*}{FC-CLIP} 
  & \multirow{4}{*}{ConvNeXt-L} & COCO & 26.8 & 16.8 & 34.1 &14.8&18.2&58.4&81.8\\
  &  & COCONut-S & 27.3 & 17.3 & 33.8 & 15.3 & 20.4 & 57.5 & 82.1\\
  &  & COCONut-B & 27.4 & 17.4 & 33.7 &15.5 & 20.1 & 58.5 & 82.0\\
  &  & COCONut-L & 27.5 & 17.4 & 33.9 &15.6 & 20.6 & 58.0 & 81.9\\
\end{tabular}
}
\caption{
\textbf{Benchmarking Open-Vocabulary Segmentation:}
We ablate the effect of using different training data to train the mask proposal network of FC-CLIP~\cite{yu2023convolutions}.
The performance is evaluated on multiple segmentation datasets in a zero-shot manner.
}
\label{tab:fcclip}
\end{table*}

\begin{table}[t!]
\tablestyle{2pt}{1.1}
\scalebox{0.8}{
\begin{tabular}{c|c|c||c|c|c}
 & & & COCO-val & relabeled COCO-val & COCONut-val \\
method  & backbone & training set &
$\text{AP}^{\text{box}}$  & $\text{AP}^{\text{box}}$  & $\text{AP}^{\text{box}}$  \\
\shline
\multirow{8}{*}{DETA}  & \multirow{4}{*}{ResNet50}  & COCO & 50.4 & 49.5 & 46.1 \\
 & & COCONut-S  & 47.8  & 53.8 & 49.5\\
 & & COCONut-B  & 50.4 & 54.4 & 51.4\\
 & & COCONut-L  & 50.6  & 54.9 & 53.7 \\
\cline{2-6}
 & \multirow{4}{*}{Swin-L}   & COCO    & 59.1
 & 58.6 & 56.1\\
 & & COCONut-S  & 54.5 & 61.3 &  58.9 \\
 & & COCONut-B & 59.3 & 62.2 & 60.1 \\
 & & COCONut-L  & 60.1 & 62.3 & 61.7 \\
\end{tabular}
}
\caption{
\textbf{Benchmarking Bounding Box Object Detection:}
We conduct the experiments using the DETA framework~\cite{ouyang2022nms}, employing various backbones and diverse training and validation sets.
The backbones are exclusively pretrained on ImageNet~\cite{russakovsky2015imagenet}. 
}
\label{tab:benchmark_box}
\end{table}

\begin{table}[t!]
\tablestyle{2pt}{1.1}
\scalebox{0.9}{
\begin{tabular}{l|l||cc|cc|cc}
& & \multicolumn{2}{c|}{COCO-val}  & \multicolumn{2}{c|}{relabeled COCO-val} & \multicolumn{2}{c}{COCONut-val} \\
method & training set & FID $\downarrow$
  & mIoU $\uparrow$ & FID $\downarrow$
  & mIoU $\uparrow$ & FID $\downarrow$
  & mIoU $\uparrow$\\
\shline
\multirow{2}{*}{GLIGEN}  &COCO         &  18.51   &        32.1        &  -           & 33.7 & 17.4 & 30.9 \\
&COCONut-S    &  18.39    &     30.4          &   -&   34.8       &  16.8 & 32.6\\
\end{tabular}

}

\caption{
\textbf{Benchmarking Mask-Conditional Image Synthesis:}
We conduct the experiments using the GLIGEN framework~\cite{li2023gligen}
mIoU is measured with another off-the-shelf  Mask2Former~\cite{cheng2022masked}, as a referee. 
}
\label{tab:gligen}
\end{table}

\textbf{Instance Segmentation:} We benchmark kMaX-DeepLab on the task of instance segmentation. Different from Tab.~6 in the main paper where the mask AP is evaluated using a panoptic segmentation model, we train a task-specific model on instance segmentation with instance masks only.
\tabref{tab:benchmark_inst_seg} summarizes the results. 
Similar to the findings in panoptic segmentation, we observe consistent improvements across various backbones (ResNet50~\cite{he2016deep} and ConvNeXt-L~\cite{liu2022convnet}.
Additionally, as we increase the size of training dataset, we observe that the improvement gain is decreasing while evaluated on the small COCO-val and relabeled COCO-val set, indicating the performance saturation on the small validation set.
By contrast, the proposed COCONut-val set presents a more challenging validation set, where the improvement of stronger backbone and more training images are more noticeable.

\textbf{Semantic Segmentation:}
We also conduct experiments on training a single semantic segmentation model with semantic masks. Results are shown in ~\tabref{tab:benchmark_sem_seg}. Similar observations are made. We can see subsequent mIoU gains of increasing the training dataset size for semantic specific model. Additionally, we also verify the dataset on semantic segmentation using ViT backbones~\cite{dosovitskiy2021vit}.
We follow the same configuration and use the codebase from ViT-Adapter~\cite{chen2022vitadapter} to conduct our experiments but replace the dataset from COCO-stuff to our COCONut semantic segmentation dataset. As shown in~\tabref{tab:rebuttal_vitadapter}, a similar observation is made: the model saturates when testing on our relabeled COCO-val set but the performance is improved on COCONut-val.

\textbf{Open-Vocabulary Segmentation:} \tabref{tab:fcclip} summarizes the results for open-vocabulary segmentation using FC-CLIP.
As shown in the table, FC-CLIP benefits from COCONut's high-quality and large-scale annotations, achieving the performance of 27.5 PQ on ADE20K, setting a new state-of-the-art.

\textbf{Bounding Box Object Detection:}
The results for object detection are shown in ~\tabref{tab:benchmark_box}. As shown in the table, the detection model with ResNet50 benefits significantly from the high quality data COCONut-S with a large margin of 4.3 on relabeled COCO-val set. Similarly, subsequent gains from training size are observed for both backbones.

\textbf{Mask Conditional Image Synthesis:} We conduct mask conditional image synthesis to verify the annotation quality for generation. We employ a mask-conditional model GLIGEN~\cite{li2023gligen} and train the model on paired image-mask data from COCO and COCONut-S separately. Once we have the trained model checkpoint, we perform inference on mask-conditioned generation by giving masks from  COCO val set, relabeled COCO-val set, and COCONut-val  set individually to compute FID. The lower FID shows better image synthesis performance. Besides, we adopt the off-the-shelf Mask2Former~\cite{cheng2022masked} model to perform semantic segmentation by giving the generated images as input and report mIoU for evaluation. As shown in Tab.~\ref{tab:gligen}, our high-quality mask annotation can result in better image synthesis with 18.39 FID on COCO-val set and 16.8 FID on COCONut-val set. Besides, the higher-quality generated images can be better inferred via the higher segmentation mIoU scores. Even for a more challenging val set, training on COCONut-S outperforms the COCO dataset.

\section{Label Map Details and Annotation Instruction}
\label{sec:label_map}
\begin{table*}[ht!]
  \centering
  \begin{tabularx}{\textwidth}{p{1cm}| p{2cm}|p{4.5cm}|X}
   
     & \textbf{category} & \textbf{COCO definition} &  \textbf{COCONut definition}\\
    \hline
    \shline
    `thing' & bed & None & A piece of furniture for sleep or rest, typically a framework with a mattress and coverings (from Google dictionary). Thus we will include the pillows, comforter, blanket, and bedding sheets along with the bed frame for bed.\\
    
    \hline
   `stuff'  & blanket & A loosely woven fabric, used for warmth while sleeping. & As blanket in the bed is included in the category of bed, then we will label blanket on the other surface excluding bed, for example, blanket on the couch or blanket on the bench. \\
   \cline{1-4}
    `stuff' & pillow & A rectangular cloth bag stuffed with soft materials to support the head. & To avoid the conflicts from bed in `thing', we exclude the pillow in the bed while labeling pillow.\\
      \hline
    \hline
     `thing' & dining table &  None & A table on which meals are served in a dining room (from Google dictionary). In order to have consistent COCO's definition by viewing hundreds of examples, partial table placing with food is also considered as dining table.  \\
 \cline{1-4}
      `stuff' & table-merged & A piece of furniture with a flat top and one or more legs. & Exclude the cases from dining table aforementioned. Include console table, coffee table, desk and \etc. \\
    
    \hline
    \hline
    `stuff' & roof  & The structure forming the upper covering of a building.   & The structure forming the upper covering of a building or vehicle (from Google dictionary). Only the outside coverings will be labeled.  COCO also labels the inner side of the coverings while they should be referred as ceiling instead. \\
    \cline{1-4}
    `stuff' & house & A smaller size building for human habitation.  & A building for human habitation, especially one that is lived in by a family or small group of people (from Google dictionary). Typically it refers to residential house meanwhile residential apartment building is not included. To avoid overlap from roof, a house will not to separated into the parts of roof and the remaining while this happens in COCO. \\
    \cline{1-4}
    `stuff' & building-other-merged & Any other type of building or structures. & For the other types of buildings, it consists of diverse types of constructions, for example, churches, stadiums, and \etc .\\

    \hline
    \hline
     `stuff' & wall-tile & A building wall made of tiles, such as used in bathrooms and kitchens.  &  Follow the same definition from COCO. \\
     \cline{1-4}
     `stuff' & wall-stone & A building wall made of stone. & Indoor wall with specific texture of stone and partial outside wall of the building instead of the whole building. In other word, the building built with stone will be labeled as building instead of wall-stone. \\
     \cline{1-4}
     `stuff' & wall-wood & A building wall made of wooden material. & Indoor and outside wall made of wood instead of the whole building. \\
      \cline{1-4}
     `stuff' & wall-brick & A building wall made of bricks of clay. &Indoor and outside wall made of bricks instead of the whole building.  \\
     \cline{1-4}
   `stuff' & wall-other-merged &  Any other type of wall. & To avoid the conflicts wall categories, we will first label the categories with specific texture and at last we label wall-other-merged. In details, we only label indoor scenes for wall-other-merged, for outdoor scenes, we will use other categories. We also need to exclude the other objects hang on the wall, for example, the frames \etc. \\
    
  \end{tabularx}
  \caption{
  \textbf{Clear Redefinition of Specific COCO Categories:}
  We present the class definitions by grouping confusing categories for easier comparison to facilitate their distinction (continued in~\tabref{tab:class_def_split2}).
  }
  \label{tab:class_definitions_1}
\end{table*}

\begin{table*}[ht!]
  \centering
  \begin{tabularx}{\textwidth}{p{1cm}| p{2cm}|p{5cm}|X}
   
     & \textbf{category} & \textbf{COCO definition} &  \textbf{COCONut definition}\\
    \hline
    \shline
    `stuff' & gravel & A loose aggregation of small water-worn or pounded stones. & Follow the same definition from COCO. \\
    \cline{1-4}
     `stuff'& railroad & A track made of steel rails along which trains run (incl. the wooden beams). &  We found that railroad often consists of the gravel and the track. In this scenario, we separate the region of gravel to be labeled as gravel and the remaining parts of the track will be labeled as railroad.\\
    \hline
    \hline
    `stuff' & playingfield &  A ground marked off for various games (incl. indoor and outdoor). & Follow the same definition. But we found COCO has a large amount of missing masks for playingfield which are mislabeled as dirt-merged instead. We label all the playingfields if they can be identified no matter 
    they are grass based or dirt based grounds. \\
    \cline{1-4}
     `stuff'& dirt-merged & Soil or earth (incl. dirt path). &  Follow the same definition but exclude dirt-based playingfields. \\
        \hline
    \hline
    `stuff' & pavement-merged &  A typically raised paved path for pedestrians at the side of a road. & Follow definition from COCO, to be more concrete, it includes side walk. \\
    \cline{1-4}
    `stuff' & platform & A paved way leading from one place to another. & COCO does not have consistent labeling masks for platforms while some of them are labeled as pavement-merged. We have a unified definition to take care of these cases. In particular, we label all the paved way for transportation, for example, label the pavement area for the train, subway and \etc as platform. \\
    \hline
    \hline
    `stuff' & net & An open-meshed fabric twisted, knotted, or woven together at regular intervals. & Follow the same definition but exclude fence made by net. \\
    \cline{1-4}
    `stuff' & fence-merged & 	A thin, human-constructed barrier which separates two pieces of land. & COCO has inconsistent masks for fence-merged and net. We follow a consistent definition to distinguish net from fence when it is not used as a fence to separate two pieces of land.  \\
    \hline
    \hline
    `thing' & potted plant & None &  A plant that is grown in a container, and usually kept inside.  There exist masks for flower placed in the vase which contradicting the definition of flower and vase. We exclude these scenarios from potted plant.  \\
    \cline{1-4}
    `thing' & vase & None & A decorative container, typically made of glass or china and used as an ornament or for displaying cut flowers (google dictionary). \\
    \cline{1-4}
     `stuff' & flower & The seed-bearing part of a plant (incl. the entire flower). &COCO does not clarify that whether the flowers that are placed in the vase belong to potted plant or flower. This is confusing when our raters label the images. We give the definition to separate the flower, potted plant and vase. The potted plant will not include any plants which are flowers. Then the potted plant will be labeled together with the plants and pots. While for vase, if the vase has flower, then these parts need to be separate. \\

\end{tabularx}
\caption{
\textbf{Clear Redefinition of Specific COCO Categories:}
  We present the class definitions by grouping confusing categories for easier comparison to facilitate their distinction (continued in~\tabref{tab:class_def_split3}).
}
\label{tab:class_def_split2}
\end{table*}

\begin{table*}[ht!]
  \centering
  \begin{tabularx}{\textwidth}{p{2.5cm}|p{5cm}|X}
   
      \textbf{category} & \textbf{COCO definition} &  \textbf{COCONut definition}\\
    \hline
    \shline
      food-other-merged & Any other type of food. & To avoid the conflicts from similar categories of `thing', we explicitly highlight that we DO NOT label those categories. The categories include sandwich (burger), hot dog, pizza, donut, cake, broccoli and carrot. Excluding all the food aforementioned, the other types of food need to be labeled.\\
    \hline 
    paper-merged & A material manufactured in thin sheets from the pulp of wood. & Include tissue, toilet paper, poster, kitchen paper towel, and \etc. They are often shown with a single or multiple pieces of papers. \\
    \hline
    tree-merged & A woody plant, typically having a single trunk growing to a considerable height and bearing lateral branches at some distance from the ground. & Include bush. \\

    \hline
    fruit & The sweet and fleshy product of a tree or other plant. & Exclude fruits in `thing', banana, orange and apple. Include tomato and all other kinds of fruit.
    \end{tabularx}
     \caption{
     \textbf{Clear Redefinition of Specific COCO Categories:} We clearly redefine certain COCO categories to avoid annotation confusion.
     }
    \label{tab:class_def_split3}
\end{table*}

Our label map definition strictly follows COCO~\cite{lin2014microsoft}.
However, the COCO definitions of specific categories might be ambiguous.
Specifically, we have identified several conflicts between the `thing' and `stuff' categories, often resulting in potential mask overlaps.
To mitigate the issue, we have meticulously redefined specific categories, detailed in ~\tabref{tab:class_definitions_1}, \tabref{tab:class_def_split2}, and ~\tabref{tab:class_def_split3}.
The definitions of categories not included in the tables remain consistent with the original COCO label map.

{
    \small
    \bibliographystyle{ieeenat_fullname}
    \bibliography{main}
}

\end{document}